%% file: main.tex
\documentclass[journal]{IEEEtran}
\usepackage{times}

\usepackage[numbers, sort&compress]{natbib}
\usepackage[T1]{fontenc}
\usepackage{multicol}
\usepackage{basicmath}
\usepackage[bookmarks=true]{hyperref}
\usepackage[capitalize]{cleveref}
\usepackage{xcolor}
\usepackage{tabularx}
\usepackage{algorithm}
\usepackage[noend]{algpseudocode}
\usepackage[inkscapearea=page]{svg}
\usepackage{duckuments}
\usepackage{stfloats}
\usepackage{siunitx}
\usepackage{subcaption}
\usepackage{makecell}

\definecolor{green}{HTML}{339933}
\definecolor{orange}{HTML}{ff9900}
\definecolor{red}{HTML}{cc0000}

\usepackage{amsthm}
\usepackage{mdframed}

\newtheoremstyle{assumption}
{1em}   
{1em}   
{}      
{}      
{\bfseries} 
{.}     
{.5em}  
{}      

\theoremstyle{assumption}
\newtheorem*{assumptions}{Assumption}

\newenvironment{justification}
{\begin{mdframed}[
		linewidth=0pt,
		leftmargin=0em,
		rightmargin=0em,
		innerleftmargin=0em,
		innerrightmargin=0em,
		innertopmargin=0.5em,
		innerbottommargin=0.5em
		]}
{\end{mdframed}}

\usepackage[export]{adjustbox}
\hypersetup{pdfauthor={Acosta},
    pdftitle={},
    pdfsubject={},
    pdfkeywords={},
    pdfproducer={LaTeX},
    pdfcreator={pdfLaTeX}
}

\begin{document}
	\newcommand{\brian}[1]{{#1}}
	\newcommand{\briannew}[1]{{#1}}
	\newcommand{\blockcomment}[1]{{}}
    \setlength\arraycolsep{3pt}
    
    \title{Perceptive Mixed-Integer Footstep Control for Underactuated Bipedal Walking on Rough Terrain}

    \author{Brian Acosta and Michael Posa 
   			\thanks{This material is based upon work supported by the National Science Foundation Graduate Research Fellowship Program under Grant No. DGE-1845298. Toyota Research Institute also provided funds to support this work.}
   			\thanks{The authors are with the GRASP Laboratory, University of Pennsylvania, Philadelphia, PA 19104, USA \{bjacosta, posa\}@seas.upenn.edu}
   		}

    \maketitle

    \begin{abstract}
        Traversing rough terrain requires dynamic bipeds to stabilize themselves through foot placement without stepping in unsafe areas.  
        Planning these footsteps online is challenging given non-convexity of the safe terrain, and imperfect perception and state estimation.
        This paper addresses these challenges with a full-stack perception and control system for achieving underactuated walking on discontinuous terrain.
        First, we develop model-predictive footstep control (MPFC), a single mixed-integer quadratic program which assumes a convex polygon terrain decomposition to optimize over discrete foothold choice, footstep position, ankle torque, template dynamics, and footstep timing at over 100 Hz.
        We then propose a novel approach for generating convex polygon terrain decompositions online. 
        Our perception stack decouples safe-terrain classification from fitting planar polygons, generating a temporally consistent terrain segmentation in real time using a single CPU thread.
        We demonstrate the performance of our perception and control stack through outdoor experiments with the underactuated biped Cassie, achieving state of the art perceptive bipedal walking on discontinuous terrain. \\
        Supplemental Video: (Short~\cite{supp_vid_short}, Long~\cite{supp_vid_full}).
    \end{abstract}

    \IEEEpeerreviewmaketitle
    \input{chapters/intro}

    \input{chapters/background}
    \input{chapters/prelim}
    \input{chapters/mpfc}
    \input{chapters/osc}
    \input{chapters/perception}
    \input{chapters/setup}
    \input{chapters/results}
    \input{chapters/discussion}
    \input{chapters/conclusion}

    \section*{Acknowledgements}
    We thank the DAIR Lab for the time they have dedicated to Cassie experiments.

    \bibliographystyle{IEEEtran}
    \small
    \bibliography{references}
    
    \appendix
    \input{chapters/appendix}

\end{document}

%% file: chapters/intro.tex
\section{Introduction}\label{sec:intro}
Bipedal robots can theoretically traverse challenging terrain by breaking contact with the ground to clear obstacles, making them potentially useful for disaster response, planetary exploration, and deployment in cluttered homes.
However, dynamic bipedal walking over rough terrain remains challenging for today's perception and control algorithms.
To traverse rough terrain, bipeds must quickly identify safe footstep positions which maintain the robot's balance and make progress in the desired walking direction.
This is a highly coupled problem where online terrain estimation is used to control an underactuated hybrid system.
Despite the existence of mature techniques for both underactuated walking, and footstep planning over constrained footholds, few works attempt to address both problems at once.
Often, underactuated gaits are stabilized within a fixed sequence of stepping-stone constraints~\cite{nguyen3DDynamicWalking2016, xiangAdaptiveStepDuration2024, daiBipedalWalkingConstrained2022}, or rough terrain is assumed to have varying height but no unsafe footstep positions~\cite{krishnaLearningLinearPolicies2021, xiongSLIPWalkingRough2021}.
Without these combinatorial aspects, the optimal control problem is easier to solve, but we have an incomplete solution for walking on rough terrain.

This paper presents Model Predictive Footstep Control (MPFC), a model-predictive-control-style footstep planner which reasons over many of the relevant decision variables for underactuated walking.
In addition to discrete foothold selection, MPFC optimizes over the continuous footstep positions, center of mass trajectory, ankle torque, and gait timing.
MPFC is one of the first controllers to simultaneously optimize over the discrete choice of stepping surface and the robot's dynamics in real time \footnote{\cite{shimTopologyBasedMPCAutomatic2023} was published concurrently with the conference version of this paper~\cite{acostaBipedalWalkingConstrained2023} and uses artificial potentials to snap footsteps onto nearby footholds, and \cite{guRobustLocomotionbyLogicPerturbationResilientBipedal2024} was published shortly after \cite{acostaBipedalWalkingConstrained2023}, and enforces stepping stone constraints with offline-generated signal-temporal-logic objectives.}, and to our knowledge, this paper and its precursor \cite{acostaBipedalWalkingConstrained2023} represent the first deployment of such a controller on hardware.

\begin{figure}[t]
	\centering
	\includegraphics[width=\linewidth]{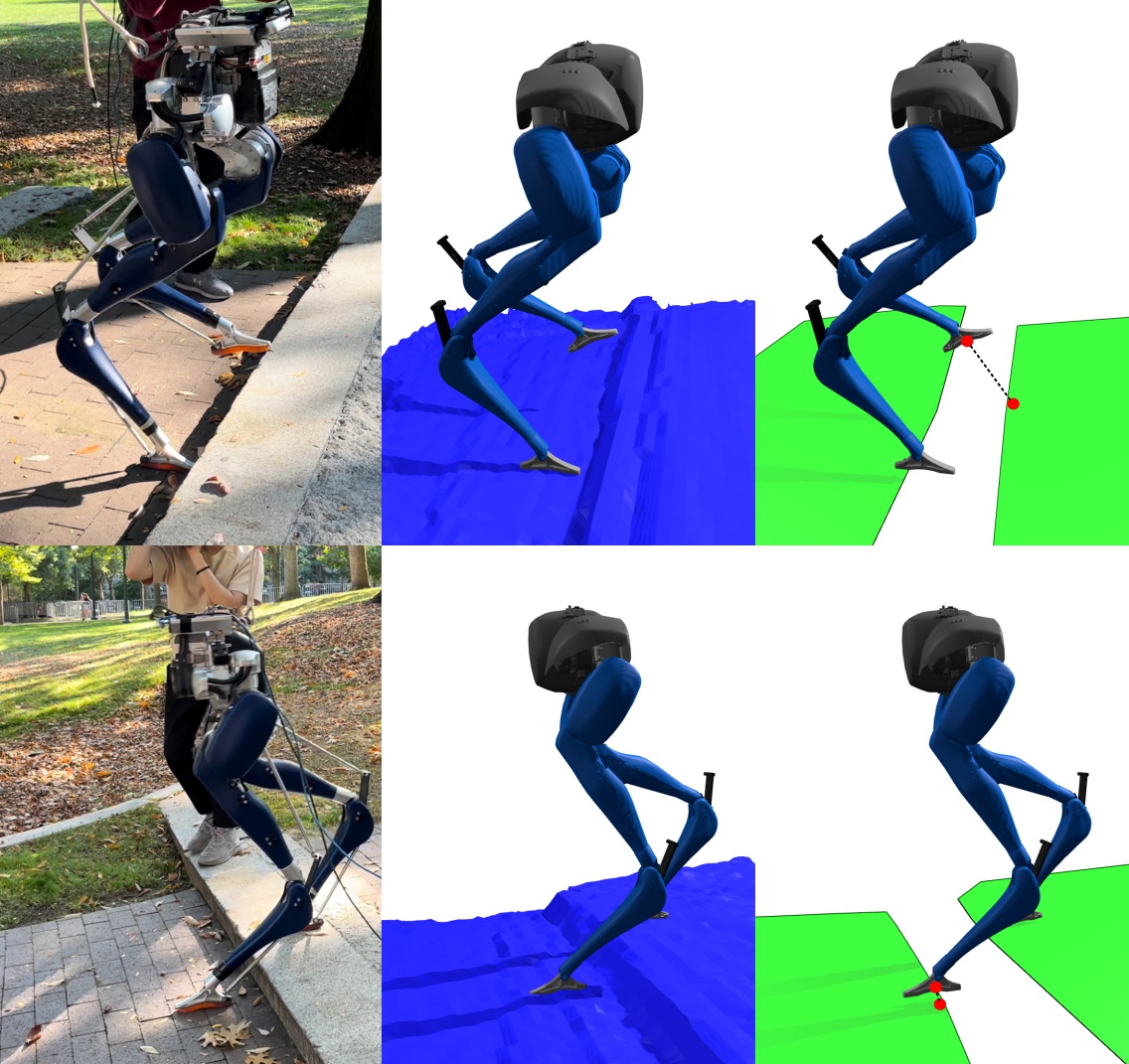}
	\caption{The bipedal robot Cassie walks up and down brick steps using the perception and control framework developed in this paper. Left: the physical robot and steps. Middle: an elevation map of the steps. Right: a convex decomposition of the safe terrain. \brian{Our MPC footstep planner constrains the center of Cassie's foot to a convex polygon foothold for each planned footstep. These convex footholds are generated online via Stable Steppability Segmentation, our novel terrain segmentation approach  designed for temporal consistency of the safe terrain classification.}
	\vspace{-1em}}
	\label{fig:hero}
\end{figure}

We use binary variables to assign \brian{the center of} each footstep to a convex foothold~\cite{deitsFootstepPlanningUneven2014}, providing a straightforward extension of linear-quadratic MPC footstep controllers~\cite{gibsonTerrainAdaptiveALIPBasedBipedal2022} to discontinuous terrain, with the consequence that optimal control problem graduates in difficulty from a Quadratic Program to a Mixed-Integer-Quadratic Program (MIQP).
MIQPs have been used extensively for offline trajectory optimization over broken terrains~\cite{aceituno-cabezasSimultaneousContactGait2018, fey3DHoppingDiscontinuous2024, dingKinodynamicMotionPlanning2020}, but due to their combinatorial complexity in the planning horizon, they have seen much less use in real-time control.
Our controller achieves solve times of less than 10 milliseconds by using a low dimensional, linear dynamics model, planning over a short footstep horizon, and eliminating foothold candidates far from the robot.

\begin{figure*}[t]
	\centering
	\includegraphics[width=0.95\textwidth]{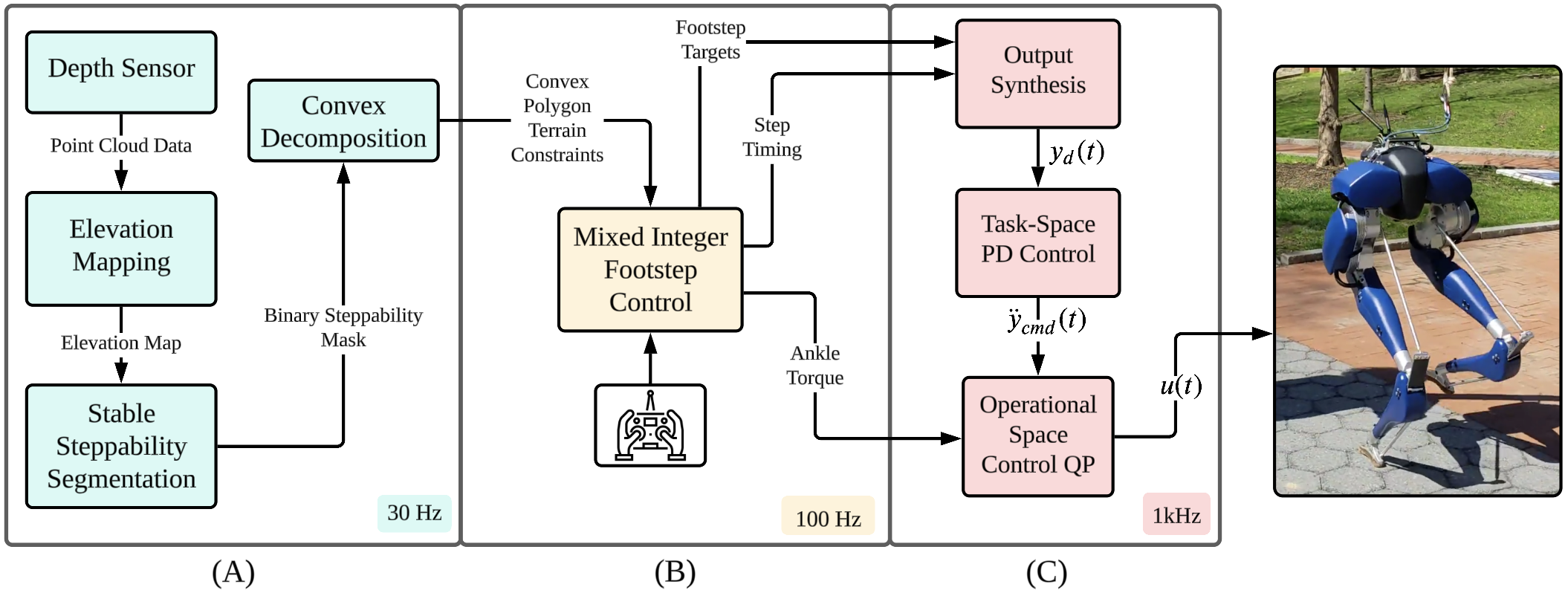}
	\caption{The perception and control stack proposed in this paper to achieve underactuated walking over discontinuous terrain. 
		Our perception stack (A) generates convex polygon foothold constraints for MPFC, a mixed-integer MPC style footstep planner (B).
		MPFC sends the next footstep, step timing adaptation, and ankle torque plan to a low-level operational-space-control process (C) which performs kHz level torque control.}
	\label{fig:block_diagram}
\end{figure*}

A significant barrier to deploying mixed-integer footstep planning methods on hardware is the need for a convex planar polygon decomposition of the terrain around the robot. 
As we discovered during the hardware experiments for our original mixed-integer footstep planning work \cite{acostaBipedalWalkingConstrained2023}, and as others have noted in recent literature \cite{corberesPerceptiveLocomotionWholeBody2023}, explicit plane segmentation approaches suffer from poor temporal consistency. 
This destabilizes online footstep planners with constraints that "flicker" into and out-of existence. 

We propose a new approach to terrain segmentation and convex decomposition. 
We argue that requiring a one-to-one correspondence between foothold constraints in the controller and real planar polygons in the environment~\cite{grandiaPerceptiveLocomotionNonlinear2022, corberesPerceptiveLocomotionWholeBody2023} is overly restrictive and brittle. 
Our approach recognizes that planar polygons are a modeling choice used to support optimization based control, rather than a hard safety requirement. 
By focusing on avoiding terrain which is clearly unsafe, we arrive at a simple algorithm which is robust to non-planar surfaces and more temporally consistent than explicit plane segmentation.

Our perception stack consists of two stages. 
The first stage, called Stable Steppability Segmentation (S3), uses local safety criteria and a simple hysteresis mechanism to classify elevation map pixels as safe or unsafe, resulting in a binary steppability mask.  
The second stage of our segmentation algorithm generates a set of convex polygons approximating the safe terrain identified by S3.
We perform approximate convex decomposition\cite{lienApproximateConvexDecomposition2006}, then take a convex inner-approximation of the resulting polygons before finally fitting plane parameters to these convex polygons using the original elevation map. 
While the S3 implementation in this paper uses intuitive heuristic criteria for steppability classification, the general algorithm supports any number of criteria, allowing for composition with learning-based approaches and higher-level obstacle detectors.

An earlier version of MPFC was presented in \cite{acostaBipedalWalkingConstrained2023}, with limited hardware results due to the brittleness of plane segmentation. 
This article extends that work by introducing our new terrain segmentation approach, improving our MPFC formulation, and presenting hardware experiments that demonstrate the capabilities of our perception and control stack.

The primary contributions of this paper are:

\begin{enumerate}
    \item We present a new terrain segmentation framework which is faster and more temporally consistent than explicit plane segmentation.
    \item We propose a new MPFC formulation which jointly optimizes over the robot's discrete choice of stepping surface, footstep plan, ankle torque, and step duration, using the step-to-step ALIP dynamics.
    \brian{Compared to~\cite{acostaBipedalWalkingConstrained2023}, we include optimization over the initial stance duration, improving the system's ability to walk over complex terrains.
    We reduce the number of MPFC variables by restricting ankle torque to the initial single-stance phase, and using a step-to-step dynamics approximation for the subsequent stance phases.
    This variable count reduction results in shorter solve times.}    
\item The resulting full-stack system is validated with hardware experiments that demonstrate real--time perceptive, dynamic, underactuated walking over constrained footholds.
\end{enumerate}

%% file: chapters/background.tex
\section{Related Work}\label{sec:background}
Our controller and perception stack build on several mature or maturing techniques such as mixed--integer--convex footstep planning, (A)LIP based footstep control, and elevation mapping for mobile-robot navigation. 
We review these topics, focusing on features of our approach compared to what exists in the literature. 

\subsection{Footstep Planning over Rough Terrain}
The literature on safe bipedal footstep planning mainly considers humanoid robots with large feet~\cite{calvertFastAutonomousBipedal2022}, which allow a feasible center of mass trajectory to be planned and tracked for any reasonable footstep plan.
These plans can be generated quickly via motion planning approaches like graph search~\cite{griffinFootstepPlanningAutonomous2019a} or mixed--integer--convex programming~\cite{deitsFootstepPlanningUneven2014}.
However, because footsteps are not re-planned at high rates, robots using decoupled approaches walk slowly to avoid violating zero-moment-point constraints~\cite{kajitaBipedalWalking2003}.

\subsubsection{Mixed-Integer Footstep Planning}
Deits and Tedrake introduced the use of MIQPs for footstep planning in~\cite{deitsFootstepPlanningUneven2014} by decomposing safe terrain into a collection of convex polygons, and using integer variables to assign every footstep to a polygon.
Tonneau et al. \cite{tonneauSL1MSparseL1norm2020} provide a convex approximation of this problem as a linear program, and Song et al. \cite{songSolvingFootstepPlanning2021} show how these approaches can be made more efficient by using a simplified trajectory planner to prune irrelevant footholds.
In contrast to our work, these works focus on long horizon footstep planning, and only consider geometric criteria such as workspace constraints, and quasistatic stability criteria, such as the existence of a feasible center of mass trajectory which lies completely above the support polygon.

MIQP footstep planning has also been used for quadruped robots.
In~\cite{risbourgRealtimeFootstepPlanning2022}, Risbourg et al. use the convex relaxation from~\cite{tonneauSL1MSparseL1norm2020} online to project the desired footstep sequence to the closest convex footholds, subject to kinematic constraints.
In~\cite{corberesPerceptiveLocomotionWholeBody2023}, Corberes et al. incorporate this footstep planning strategy as an online foothold scheduler at 1-5 Hz with vision in the loop.
Due to the low planning rate, and the lack of dynamics constraints in the contact scheduler, they rely on a separate whole body MPC to find feasible robot trajectories.
Aceituno-Cabezas et al. \cite{aceituno-cabezasSimultaneousContactGait2018} formulate a full quadruped trajectory optimization problem using mixed integer constraints for footholds and to approximate the nonlinear manifold constraint for 3D rotations as piecewise linear.
Their trajectory optimization features both kinematic and dynamics constraints, but does not re-plan the footholds in real time.

\subsection{Footstep Control for Underactauted Bipeds}
Dynamic walking research assumes minimal ankle actuation, instead viewing walking as controlled falling, where momentum can only be added or removed from the system by stepping to the appropriate spot on the ground.
These approaches generally assume flat or constantly sloped ground without obstacles to synthesize reactive stepping controllers based on the linear inverted pendulum~\cite{kajita3DLinearInverted2001a, xiong3DUnderactuatedBipedal2022, griffinReachabilityAwareCapture2023}.
This approach regulates walking speed without ankle torque by using foot placement to affect the initial conditions of each single stance phase.
Combined with output tracking via inverse-dynamics based whole body torque controllers, this approach has enabled dynamic and robust walking.
The Angular Momentum Linear Inverted Pendulum (ALIP) model, in particular, has been shown to accurately describe the bulk motion of walking even for robots with heavy legs \cite{gongOneStepAheadPrediction2021}, and has been used to stabilize walking on sloped terrain \cite{gibsonTerrainAdaptiveALIPBasedBipedal2022}, synthesize specialized stair climbing controllers \cite{dosunmu-ogunbiStairClimbingUsing2023}, and walk on pre-selected constrained footholds \cite{daiBipedalWalkingConstrained2022}.

\subsection{Safe Terrain Estimation for Legged Locomotion}
Elevation maps are a convenient intermediate terrain representation for legged locomotion due to their ability to fuse multiple sensor streams over time in a compact representation \cite{fallonContinuousHumanoidLocomotion2015a, fankhauserProbabilisticTerrainMapping2018, mikiElevationMappingLocomotion2022}. 
This has lead to a proliferation of algorithms for extracting convex planar polygons from elevation maps via plane segmentation \cite{mikiElevationMappingLocomotion2022, fallon_plane_2019}.
However, these approaches segment each elevation map independently, leading to issues with temporal consistency~\cite{corberesPerceptiveLocomotionWholeBody2023, acostaBipedalWalkingConstrained2023}, especially because elevation mapping is vulnerable to artifacts from drift in the floating base position estimate \cite{wisthVILENS2023}.
To generate temporally consistent polygon constraints in real-time, despite these challenges imposed by legged locomotion,  Bin et al. develop a GPU accelerated semantic mapping framework \cite{binRealTimePolygonal2024} to directly estimate the state of polygonal terrain from depth images. 
This approach has advantages for stair climbing, where the terrain is known to be planar and precise foot placement is required, but could struggle in outdoor environments where the ground is not perfectly flat. 

On unstructured terrain, some works compute heuristic costs from the elevation map to guide planning. 
McRory et al. encode various traversability costs into a graph search algorithm for humanoid footstep planning  \cite{mcroryBipedalNavigation2023}.
Jenelten et al.  add a nonconvex cost on the gradient of the elevation map at the planned stance foot locations in their MPC formulation for quadrupedal walking \cite{jeneltenTAMOLSTerrainAwareMotion2022}.  
These heuristic costs recognize that planar polygons are a modeling choice to support optimization based control, not a necessary condition for steppability. 
Our terrain segmentation approach adopts a similar philosophy by classifying elevation map cells as steppable or not without regard for global planarity, but still achieves global optimality in the MPC problem by transforming this classification back into mixed-integer convex terrain constraints. 

\subsection{Reinforcement Learning for Legged Locomotion}
Sim-to-Real reinforcement learning (RL), where control policies are learned in simulation and then deployed on hardware, has seen increasing success in recent years, especially for legged locomotion. 
These policies can be made robust and performant through a combination of domain randomization and adaptation. 
For example, Siekmann et al. learn a blind stair climbing controller for Cassie in \cite{siekmannBlindBipedalStair2021}, and Duan et al. use a similar policy to walk on constrained footholds in \cite{duanSimtoRealLearningFootstepConstrained2022}.
With additional vision modules, they achieve perceptive locomotion over boxy terrain as well \cite{duan2023learningVisionBased}.
Because RL can struggle with sparse footholds, Jenelten et al. proposed a hierarchical approach where a model-based footstep planner guides a lower level RL tracking policy \cite{jeneltenDTC2024}.
Yu et al. propose the opposite, where RL learns high level strategies like gait selection and foot placement, and MPC generates and stabilizes corresponding full body motions \cite{yu2024LearningGeneric}.
This strategy is now controlling Boston Dynamics' Spot quadruped in industrial use cases~\cite{bostonDynamicsSpotRL}.
While RL is not a focus of this paper, the subcomponents of our stack (including our segmentation module) could be used in one of these hierarchical frameworks. 

%% file: chapters/prelim.tex
\section{Preliminaries}\label{sec:prelim}
This section overviews the reduced-order Angular-Momentum Linear Inverted Pendulum (ALIP) model used in MPFC, and the operational space controller used to track MPFC's outputs. 
We start by reviewing the ALIP dynamics, then we derive the reset map and step-to-step dynamics for a hybrid ALIP model with a finite double stance period. 
We then provide a linearization of solutions to the ALIP model with respect to time be used for stance timing adaptation.
Finally we overview our inverse-dynamics operational space controller for output tracking based on MPFC solutions. 

\subsection{ALIP model}
\begin{figure}[t]
	\centering
	\includegraphics[width=0.5\linewidth]{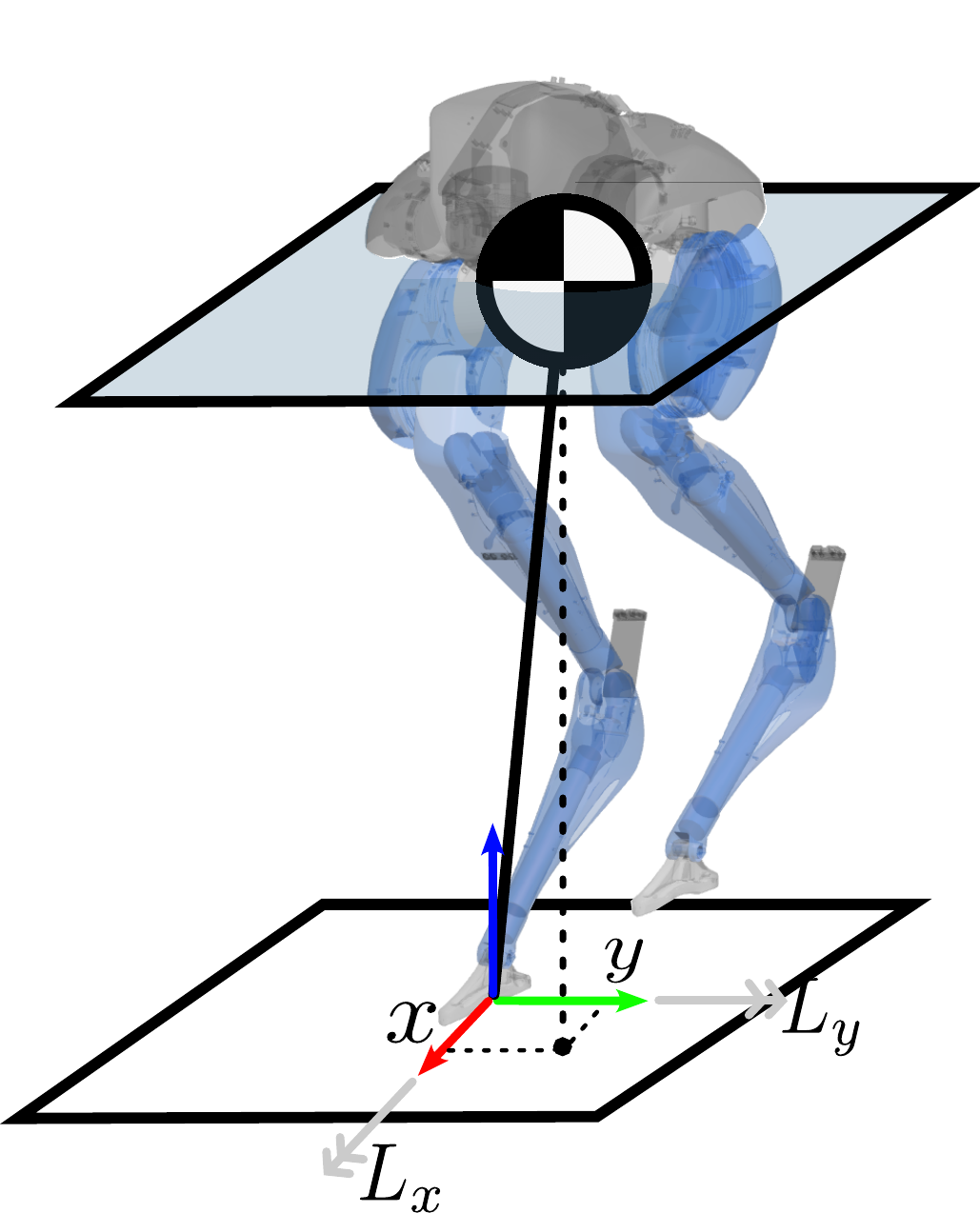}
	\caption{The ALIP model assumes that the robot's CoM is restricted to a virtual plane above the terrain. The states of the ALIP model are the horizontal CoM positions, and the angular momentum of the robot about the horizontal axes.}
	\label{fig:cassie_alip}
\end{figure}

The ALIP model (\cref{fig:cassie_alip}) is an approximation of the horizontal center-of-mass dynamics of the robot during single stance.
The ALIP model is similar to Kajita's Linear Inverted Pendulum model~\cite{kajita3DLinearInverted2001a}, but uses angular momentum about the contact point in place of center-of-mass velocity to describe the speed of the robot.
Angular momentum about the contact point has the advantage of being relative-degree three to (non-stance ankle) motor torques, compared to relative-degree one  for center-of-mass velocity~\cite{gongOneStepAheadPrediction2021}, making the predictions of the ALIP model relatively accurate even for robots with heavy legs. 
We direct the reader to~\cite{gibsonTerrainAdaptiveALIPBasedBipedal2022} for a derivation of the ALIP dynamics assuming piece-wise planar terrain with a passive ankle.
The state of the ALIP model consists of the horizontal position of the center of mass realative to the stance foot, $(x_{com}, y_{com})$ and the tilting components of the angular momentum of the robot about the contact point $(L_{x}, L_{y})$.

To take full advantage of Cassie's blade foot, we include ankle torque in the sagittal plane, $u$ as an input to the continuous time ALIP model.
The dynamics of the ALIP with ankle torque are given by
\begin{equation}
	\underbrace{\begin{bmatrix} \dot{x}_{com} \\ \dot{y}_{com} \\ \dot{L}_{x} \\ \dot{L}_{y} \end{bmatrix}}_{\dot{x}} =
	\underbrace{\begin{bmatrix} 0 & 0 & 0 & \frac{1}{mH}\\
			0 & 0 & \frac{-1}{mH} & 0\\
			0 & -mg & 0 & 0\\
			mg & 0 & 0 & 0 \end{bmatrix}}_{A} \underbrace{ \begin{bmatrix} x_{com} \\ y_{com} \\ L_{x} \\ L_{y} \end{bmatrix}}_{x} + \underbrace{\begin{bmatrix} 0\\0\\0\\1 \end{bmatrix}}_{B} u \label{eq:alip}
\end{equation}
where $m$ is the robot's mass, and $H$ is the height of the CoM above the terrain, and all quantities are in the stance frame.

\subsection{Hybrid ALIP Model Based on Foot Placement}
To enable control of the ALIP through foot placement, we derive a reset map relating the positions of the robot's feet at touchdown to a discrete jump in the ALIP state.
Many walking controllers feature a double stance phase during which weight transfers from one leg to the other.
A double stance phase is particularly useful for Cassie, to avoid oscillations caused by rapidly unloading Cassie's leaf springs.
To treat the single and double stance phases as a single step in the step-to-step dynamics \cite{xiong3DUnderactuatedBipedal2022}, we derive a reset map from $x_{-}$, the ALIP state just before footfall, to $x_{+}$, the ALIP state just after liftoff, including a double stance phase of fixed duration, $T_{ds}$.
We start by integrating the double stance dynamics, and then we apply a coordinate change to express the ALIP state with respect to the new stance foot.

During double stance, we leave the ankles passive and treat the center of pressure (CoP) between the two feet as a control input. We then integrate the resulting dynamics with an assumed input trajectory,
\begin{align}
	p_{CoP}(t) = p_{-} + f(t)(p_{+} - p_{-})
	\label{eq:cop_input}
\end{align}
where $p_{-}, p_{+} \in \mathbb{R}^{3}$ are the pre- and post-touchdown stance foot positions, $t$ is the time since the beginning of double stance, and $f(t) : \mathbb{R} \mapsto [0, 1]$ determines the rate at which weight is transferred to the new stance foot. 
The CoP enters the ALIP dynamics via the angular momentum transfer formula
\begin{equation*}
	L_{CoP} = L_{p_{-}} +  (p_{CoP} - p_{-}) \times m v_{CoM}.
\end{equation*}
To formulate the double-stance dynamics as a linear system, we introduce an assumption to remove the cross-product term:

\begin{assumptions}
	$\left[(p_{CoP} - p_{-}) \times m v_{CoM} \right]_{x, y} \approx 0.$
	\begin{justification}
		\textbf{Justification.}
		The most straightforward justification is that the robot is generally stepping in the direction it is walking, so $v_{CoM}$ is approximately parallel to $p_{+} - p_{-}$.
		We can also consider that under the ALIP model, $p_{CoP} - p_{-}$ and $v_{CoM}$ both lie in the ground plane, so  $(p_{CoP} - p_{-}) \times v_{CoM}$ must be normal to this plane. For flat ground, this vector is perpendicular to the world $x-y$ axes, and the $x$ and $y$ components remain small for non trivial slopes (e.g. $\sin(\ang{15}) \approx 0.25$). 
	\end{justification}
\end{assumptions}
Under this assumption, the tilting angular momenta about $p_{-}$ and $p_{CoP}$ are equal. This allows us to treat the CoP as a virtual contact point, yielding
\begin{align}
	\dot{L}_{x} = -mg(y_{com} - p_{CoP, y}(t)) \nonumber\\
	\dot{L}_{y} = mg(x_{com} - p_{CoP, x}(t)).
	\label{eq:alip_vcp}
\end{align} 
Substituting \eqref{eq:cop_input} into \eqref{eq:alip_vcp}, we arrive at the continuous dynamics describing the ALIP during double stance:
\begin{equation}
	\dot{x} = Ax + \underbrace{
		\begin{bmatrix} 0_{2 \times 1} & 0_{2 \times 1} & 0_{2 \times 1} \\ 0 & mg & 0 \\ -mg & 0 & 0 \end{bmatrix}}_{B_{CoP}} f(t) (p_{+} - p_{-})
	\label{eq:reset_cont}
\end{equation}

The solution to \eqref{eq:reset_cont} with the initial condition $x(0) = x_{-}$ is linear in $x_{-}, p_{-}$ and $p_{+}$ \cite{ogataModernControlEngineering}: 
\begin{equation}
	x(T_{ds}) = A_{r}x_{-} + B_{ds} \left(p_{+} - p_{-}\right).
	\label{eq:ds_reset}
\end{equation}
where $A_{r} = \exp(AT_{ds})$ and
\begin{equation}
	B_{ds} = \left(\int_{0}^{T_{ds}}f(t) e^{A(T_{ds} - t)}dt \right) B_{CoP}.
	\label{eq:b_ds_general}
\end{equation} 

For $f(t) = \frac{t}{T_{ds}}$, i.e. linearly shifting the robot's weight between feet over double stance\footnote{
Because $B_{ds}$ is decoupled in $x$ and $y$, our \brian{hardware} MPFC implementation assumes $f(t) = 1$ for the lateral compenents of the ALIP state, which corresponds to instantaneous weight transfer at the beginning of double-stance.
We detail how this helps Cassie track the desired step width in \cref{sec:reset_appendix}.}, \eqref{eq:b_ds_general} evaluates to 
\begin{equation}
	B_{ds} = A_{r}A^{-1} \left(\frac{1}{T_{ds}} A^{-1} \left(I - A_{r}^{-1}\right)  -A_{r}^{-1}\right)B_{CoP}.
	\label{eq:b_ds_t}
\end{equation} 

The remainder of the reset map is a coordinate change to express the CoM position as relative to the new stance foot,
\begin{equation}
	x_{+} = x(T_{ds}) + \underbrace{\begin{bmatrix} -I_{2\times 2} & 0_{2\times 1} \\  0_{2\times 2} &  0_{2\times 1}\end{bmatrix}}_{B_{fp}} \left(p_{+} - p_{-} \right). \label{eq:alip_reset} 
\end{equation}
with the $fp$ subscript denoting ``foot placement".

By sequentially applying \eqref{eq:ds_reset} then \eqref{eq:alip_reset}, we arrive at a reset map from $x_{-}$ to $x_{+}$ which is linear in $x_{-}, x_{+}, p_{-} \text{ and } p_{+}$, 

\begin{equation}
	x_{+} = \begin{bmatrix} A_{r} & \left(-B_{ds} - B_{fp}\right) & \smash[b]{\underbrace{\left(B_{ds} + B_{fp}\right)}_{B_{r}}}\end{bmatrix} \begin{bmatrix} x_{-} \\ p_{-} \\ p_{+} \end{bmatrix}.
	\label{eq:full_reset}
\end{equation}

\subsection{Step-to-Step ALIP Dynamics}
We will also consider step-to-step (s2s) ALIP dynamics. 
We view the ALIP without ankle actuation as a discrete-time linear time-invariant system by sampling the ALIP state at the end of each (fixed duration of $T_{ss}$) single stance phase. 
These dynamics are simply 
\begin{equation}
	x_{n+1} = A_{s2s} x_{n} + B_{s2s} (p_{n+1} - p_{n})
	\label{eq:alip_s2s}
\end{equation}
where $A_{s2s} = \exp(A(T_{ss} + T_{ds}))$ and $B_{s2s} = \exp{(A T_{ss})} B_{r}$. 

\subsection{Step Timing Adaptation}
We will use the fact that the initial ALIP state is constant to adapt the duration of the initial swing phase as part of the MPFC problem formulation. 
This has previously been applied to controllers based on the divergent component of motion \cite{englsberger2013DCM, khadivWalkingControlBased2020, xiangAdaptiveStepDuration2024} and instantaneous capture point \cite{griffinReachabilityAwareCapture2023}, as these models admit an exact coordinate transform for the initial stance duration to make the touchdown state linear in the transformed variable.  
The ALIP state space does not admit this coordinate change, so we instead linearize the solution to \eqref{eq:alip}. 
\brian{Given $T$ seconds remaining in single stance}, and a current ALIP state $x_{c}$, the exact solution to \eqref{eq:alip} with constant ankle torque, $u$, is
\begin{equation}
	x(T) = A_{d}(T) x_{c} + B_{d}(T)u
	\label{eq:xT_sol}
\end{equation}
Where $A_{d}(T) = \exp(AT)$ and $B_{d}(T) =  A^{-1}(A_{d}(T) - I)B$. 
We linearize \eqref{eq:xT_sol} with respect to $T$ and $u$ about a nominal remaining stance time of $T^{*}$ and ankle torque of $0$ to find the ALIP state at the end of the current stance period (and the initial state of the s2s ALIP model), $x_{0}$:
\begin{align}
	x_{0} = A_{d}(T^*) x_{c} + \left. \frac{\partial A_{d}}{\partial T} \right|_{T^*} (T - T^*) x_{c} + B_{d}(T^*) u.
	\label{eq:step_timing_adapt}
\end{align}

\subsection{Operational Space Control}
We use operational-space control (OSC) to track outputs such as swing foot position and pelvis orientation, while respecting frictional contact constraints \cite{wensingGenerationDynamicHumanoid2013a}. OSC considers a full-order Lagrangian model of the robot's dynamics:
\begin{align}
	M(q) \dot{v} + C(q, v) = g(q) + Bu + J_{\lambda}^{T} \lambda
\end{align}
Where $q$ and $v$ are generalized positions and velocities, $u$ are inputs, and $\lambda$ are forces arising from contacts or other holonomic constraints.
Given a set outputs to track, $\{y_{i}\}$, we define task-space PD controllers, 
$$\ddot{y}_{i, cmd}=\ddot{y}_{i, des}+K_p(y_{i, des}-y_{i})+K_d(\dot{y}_{i, des}-\dot{y}_{i}).$$
The goal of OSC is to find dynamically feasible inputs, generalized accelerations, contact forces, and constraint forces, such that the task-space accelerations, $\ddot{y}_{i} = J_{i}\dot{v} + \dot{J}_{i}v$, match the PD controller as closely as possible, while satisfying contact constraints and holonomic constraints.
We formulate this as a quadratic program with Lorentz cone constraints on the contact forces:
\begin{subequations}
	\begin{align}
		\underset{\dot{v}, u, \lambda_{h}, \lambda_{c}, \varepsilon}{\text{minimize }} & \sum_i^N\widetilde{\ddot{y}}_{i}^TW_i\widetilde{\ddot{y}}_{i}+\left\|u\right\|_W^2+\left\|\dot{v}\right\|_W^2 + \left\|\varepsilon\right\|_W^2\\
		\text{subject to } & M\dot{v} + C = g + Bu + J_{h}^{T}\lambda_{h} + J_{c}^{T}\lambda_{c}\\
		& J_{h}\dot{v} = -\dot{J_{h}}v\\
		& J_{c}\dot{v} + \varepsilon = -\dot{J_{c}}v\\
		& \lambda_{c} \in \mathcal{F}\\ 
		& u_{min} \leq u \leq u_{max}
	\end{align}
\end{subequations} 

where $\lambda_{c}$ and $J_{c}$ are the stacked contact forces and contact Jacobians, and $\mathcal{F}$ is the product of the friction cones for each contact point. The contact constraint is treated as a soft constraint by the introduction of a slack variable $\varepsilon$ to ensure the problem is always feasible. The holonomic constraint $J_{h}\dot{v} = -\dot{J}_{h}v$ represents Cassie's four-bar linkages and fixed joint constraints to model Cassie's leaf spring springs. The task space acceleration errors are $\widetilde{\ddot{y}}_{i} = \ddot{y}_{cmd} - (J_{y,i}\dot{v} + \dot{J}_{y,i}v)$. 

%% file: chapters/mpfc.tex
\section{Mixed Integer Footstep Control}\label{sec:mpfc}
This section details the formulation of our model predictive footstep controller as an MIQP (\cref{fig:block_diagram}B). 
\brian{Compared to \cite{acostaBipedalWalkingConstrained2023}, our problem statement is more expressive while using fewer decision variables.}
For the current stance phase, MPFC optimizes the stance duration and ankle torque. 
In the subsequent stance phases, MPFC only affects the s2s ALIP state through foot placement, \brian{whereas \cite{acostaBipedalWalkingConstrained2023} included several knot points per stance phase for the ALIP state and ankle torque.
Because double stance is incorporated into the s2s dynamics, MPFC merges the double and single stance phase together into a combined stance phase, which is treated as single-stance within the optimization. 
The MPFC stance phase begins at each touchdown event with a nominal remaining stance time of $T_{ss} + T_{ds}$, and the footstep and gait timing solutions are ignored until the time since touchdown exceeds $T_{ds}$ (\cref{fig:mpfc_variables}).
}

The continuous MPFC decision variables are the step-to step ALIP states, $x_{n}$, the foostep positions, $p_{n}$, a constant ankle torque during  the initial stance phase, $u$, and the \brian{remaining} duration of the current stance phase, $T$. 
We also introduce one binary variable per discrete foothold per stance phase, $\mu_{n, i}$, where $i \in 1 \ldots M$ specifies that the binary variable corresponds to $\mathcal{P}_{i}$ , one of $M$ available convex polygon footholds. 
A diagram of the key MPFC decision variables is show in \cref{fig:mpfc_variables}.

We now introduce the MPFC problem statement \eqref{eq:mpfc}, and then elaborate on the costs and constraints.
Let $x_{c}$ be the current ALIP state, with $T^*$ seconds nominally remaining in the current MPFC stance phase.
MPFC is formulated as:

\begin{subequations}
    \begin{align}
        \underset{\mathbf{x, p}, \boldsymbol{\mu}, u, T}{\text{minimize}} \text{ } & J_{mpc}(\mathbf{x, p}) +
        J_{reg}(T, u) \label{eq:mpc_cost} \\
        \text{subject to } & x_{0} =  A_{d}x_{c} + AA_d x_{c}(T - T^*) +  B_{d}u \label{eq:mpfc_timing}\\
        & x_{n+1} = A_{s2s}x_{n} + B_{s2s}(p_{n+1} - p_{n}) \label{eq:mpfc_dynamics}\\
        & \mu_{n,i} = 1 \implies p_{n} \in \mathcal{P}_{i}  \label{eq:mpc_foothold}\\
        & \sum_{i \in \mathcal{I}} \mu_{n,i} = 1 \label{eq:mpc_sum_int}\\
        & \mu_{n,i} \in \{0,1\}\label{eq:mpc_binary}\\
        & \text{CoM, Input,Timing, and Footstep limits} \nonumber
    \end{align}
    \label{eq:mpfc}
\end{subequations}
\vspace{-0.75cm}
\subsection{Cost Design}
\brian{Previous works use deviation from a reference ALIP trajectory as a state cost \cite{acostaBipedalWalkingConstrained2023, gibsonTerrainAdaptiveALIPBasedBipedal2022}.
However, this implicitly encodes the corresponding footstep sequence into the state cost, and we desire for MPFC to freely pick the appropriate footstep sequence for a given terrain.
Therefore we formulate a state cost which does not encode any particular footstep pattern.
We penalize the distance of the MPFC solution from \textit{the set} of ALIP trajectories which are periodic over 2 steps and achieve the desired velocity, $v_{des}$. 
This set is an affine subspace representing all possible $x_{n}$ which satisfy the system \eqref{eq:subspace}.

\begin{subequations}
\begin{align}
	(A_{s2s}^{2}  - I)x_{n} + A_{s2s}B_{s2s} \delta p_{n} + B_{s2s} \delta p_{n+1} = 0 \label{eq:subspace_1}\\
	\delta p_{0} + \delta p_{1} = 2T_{s2s}v_{des} \label{eq:subspace_2}
\end{align}
\label{eq:subspace}
\end{subequations}

\begin{figure}[!t]
	\centering
	\includegraphics[width=0.55\linewidth]{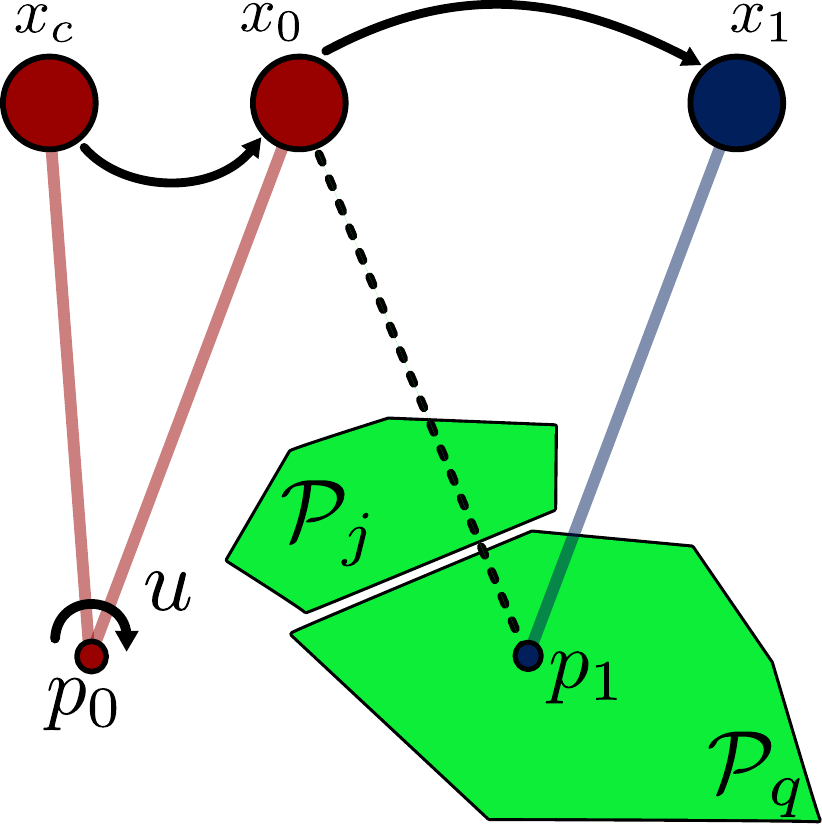}\\
	\vspace{1em}
	\includegraphics[width=\linewidth]{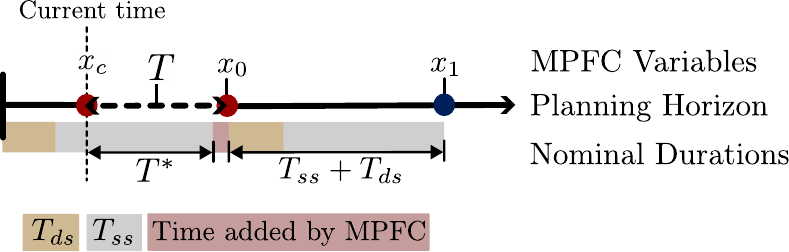}
	\caption{\textbf{Top:} Key MPFC decision variables and constraints for a horizon of 2 stance phases.
		$x_{c}$ is the current ALIP state, $u$ is ankle torque applied during the current stance phase, $x_{0}$ is the ALIP state at the end of the current stance phase, and $x_{1}$ is the ALIP state at the end of the next stance phase.
		The current stance foot position, $p_{0}$, is unconstrained, and subsequent footsteps are constrained to lie in either $\mathcal{P}_{j}$  or $\mathcal{P}_{q}$ using integer variables.
		\brian{\textbf{Bottom:} Relationship between the nominal stance phases and the MPFC gait timing optimization. The initial stance duration is adjusted continuously by optimizing over the remaining stance time, $T$.}\vspace{-0.75cm}}
	\label{fig:mpfc_variables}
\end{figure}

Where $\delta p_{n} = p_{n+1} - p_{n}$, \eqref{eq:subspace_1} is the ALIP dynamics rolled out over two footsteps, with the period-2 orbit constraint $x_{n+2} = x_{n}$, and \eqref{eq:subspace_2} requires the net displacement of the robot to match the desired velocity. 
We show how to eliminate $\delta p_{n} \text{ and } \delta p_{n+1}$ in \cref{sec:subspace}, to express solutions of \eqref{eq:subspace} as

\begin{equation}
	\Pi_{n}(x_{n}  - d_{n}(v_{des})) = 0.\\
	\label{eq:subspaces_for_cost}
\end{equation}

Where $\Pi_{n} \in \mathbb{R}^{4\times4}$ is a projection matrix used to eliminate $\delta p_{n}$ from \eqref{eq:subspace}, and $d_{n}(v_{des})$ is an offset that encodes the desired velocity. }
Our MPC cost is then formulated as 
\begin{align*} 
	J_{mpc}(\mathbf{x}, \mathbf{p}) = \sum_{n = 1}^{N-1} & \left[(x_{n} - d_{n})^{T}\Pi_{n}^{T}Q\Pi_{n}(x_{n} - d_{n}) \right. + \\
	& \left. \text{  } (\delta {p}_{n} - \delta {p}_{n}^{*})^{T}R(\delta {p}_{n} - \delta {p}_{n}^{*})\right] + \\[5pt]
	& (x_{N} - d_{N})^{T}\Pi_{N}^{T}Q_{N}\Pi_{N}(x_{N} - d_{N})
\end{align*}

where $Q, R$, and $Q_{N}$ are positive-definite weight matrices. 
We regularize the relative footstep positions to a nominal step size, defined by the desired velocity and the step width, $l$, as 

\begin{equation}
	\delta p_{n}^* = \begin{bmatrix} v_{des, x} (T_{ss} + T_{ds}) \\ v_{des, y} (T_{ss} + T_{ds}) + \sigma_{n}l \\ 0\end{bmatrix}
\end{equation}
where $\sigma_{n} = -1$ for left-stance and $+1$ for right stance. 
We add quadratic costs on $T$ and $u$, weighted by positive scalars $w_{T}$ and $w_{u}$:
\begin{equation}
		J_{reg} = w_{T}\lVert T - T^*\rVert^2 + w_{u}\lVert u \rVert^2.
\end{equation}
\vspace{-1cm}
\subsection{Dynamics Constraints}
The initial state constraint \eqref{eq:mpfc_timing} evaluates \eqref{eq:step_timing_adapt} to relate the current ALIP state to the initial s2s ALIP state via ankle torque and stance duration. 
The dynamics constraints \eqref{eq:mpfc_dynamics} are the s2s ALIP dynamics \eqref{eq:alip_s2s}.
\vspace{-0.5cm}
\subsection{Foothold Constraints}
Each convex polygonal foothold is defined by a plane $f_{i}^{T}p = b_{i}$ and a set of linear constraints $F_{i}p \leq c_{i}$.
The logical constraint~\eqref{eq:mpc_foothold} is enforced with the big-M formulation
\begin{subequations}
\begin{align}
	F_{i}p_{n} \leq c_{i} + M(1 - \mu_{n,i})\\
	f_{i}^{T}p_{n} \leq b_{i} + M(1 - \mu_{n,i})\\
	-f_{i}^{T}p_{n} \leq -b_{i}+ M(1 - \mu_{n,i}).
\end{align}
\label{eq:bigM}
\end{subequations}

With appropriately normalized $F_{i}$ and $f_{i}$, \eqref{eq:bigM} corresponds to relaxing each foothold constraint by $M$ meters when $\mu_{i} = 0$.
Since our problem scale is on the order of 2 m, we choose M = 10 for simplicity\footnote{$M$ must be large enough for every relaxed foothold to contain every unrelaxed foothold, but should otherwise be small for numerical stability}.
The binary constraint~\eqref{eq:mpc_binary} and the summation constraint~\eqref{eq:mpc_sum_int} imply that exactly one foothold must be chosen per stance phase.

\subsection{CoM, Timing, Input, and Footstep Limits}
We add the following constraints to reflect the physical limitations of the robot:
\begin{itemize}
	\item We add a soft-constraint on the CoM position of  $\pm$35 cm. in each direction.
	\item We update bounds on $T$ at each solve so the total single-stance duration lies in the range [0.27, 0.33] seconds.
	\item We add a crossover constraint to prevent the feet from crossing the $x-z$ plane.
	\item We limit the ankle torque to 22 Nm to keep the center of pressure within the blade foot.
	\item \brian{With $T_{min} = 0.27$ seconds left in the nominal single stance time, we add a trust region constraint on $p_{1}$. 
	This constraint is a bounding box centered at the previous $p_{1}$ solution with a radius of $T^{*}$ m. 
	As implied by the unit conversion of $T^*$ to a distance, the radius of this bounding box shrinks at a rate of 1 m/s.}
\end{itemize}

%% file: chapters/osc.tex
\section{Output Synthesis For Operational Space Control} \label{sec:osc}
To realize the planned walking motion on the physical robot, MPFC outputs are tracked with OSC (\cref{fig:block_diagram}C).
This section describes the construction of the outputs tracked by the OSC.

\subsection{Center of Mass Reference}\label{subsec:com_ref}
\begin{figure}[t]
	\centering
	\includegraphics[width=0.25\textwidth]{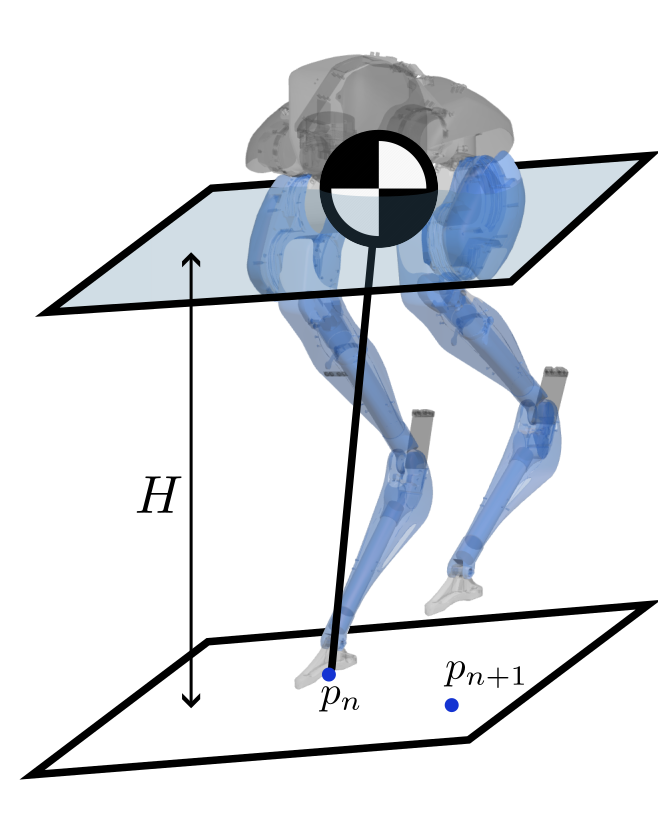}
	\caption{To enforce the planarity assumption of the ALIP, we use OSC to drive Cassie's CoM to a virtual plane defined by current and upcoming stance foot positions.}
	\label{fig:alip_plane}
\end{figure}

Given a footstep plan, we construct a CoM trajectory which enforces the local planarity assumption of the ALIP model by constructing the least-inclined plane passing through the current and imminent stance foot positions (\cref{fig:alip_plane}).
Letting $p = p_{n+1} - p_{n}$, the plane parameters are the solution to

\begin{equation}\label{eq:com_plane}
	\begin{bmatrix}
		p_{x} & p_{y}\\ -p_{y} & p_{x}
	\end{bmatrix}
	\begin{bmatrix}
	k_{x} \\ k_{y}
	\end{bmatrix} =
	\begin{bmatrix}
	p_{z} \\ 0
	\end{bmatrix}.
\end{equation}

After solving for $k_{x}$ and $k_{y}$, we define the reference trajectory for the CoM height in the stance frame as

\begin{equation}\label{eq:com_traj}
z_{c}(t) = H + k_{x} x_{c}(t) + k_{y} y_{c}(t).
\end{equation}

\brian{To account for discontinuities in $k_{x}, k_{y}, x_{c}, \text{ and } y_{c}$ when the stance foot changes, we add a first-order low pass filter on $k_{x}$ and $k_{y}$ with a 100Hz cutoff frequency, and we clip the desired $z_{c}$ to within 2.5 cm of the measured CoM height.}

\subsection{Swing Foot Reference}\label{subsec:swing_foot_ref}
We continuously adapt the swing foot trajectory $p_{sw}(t)$ to the updated swing-phase duration and planned next footstep position with a planning QP similar to \cite {khadivWalkingControlBased2020}. 
First we generate a waypoint above the line connecting the initial and final foot location, following an adaptive clearance scheme, then we find a single-segment polynomial trajectory through this waypoint. 
\subsubsection{Adaptive Swing Foot Clearance}
\begin{figure}[h!]
	\centering
	\includegraphics[width=0.66\linewidth]{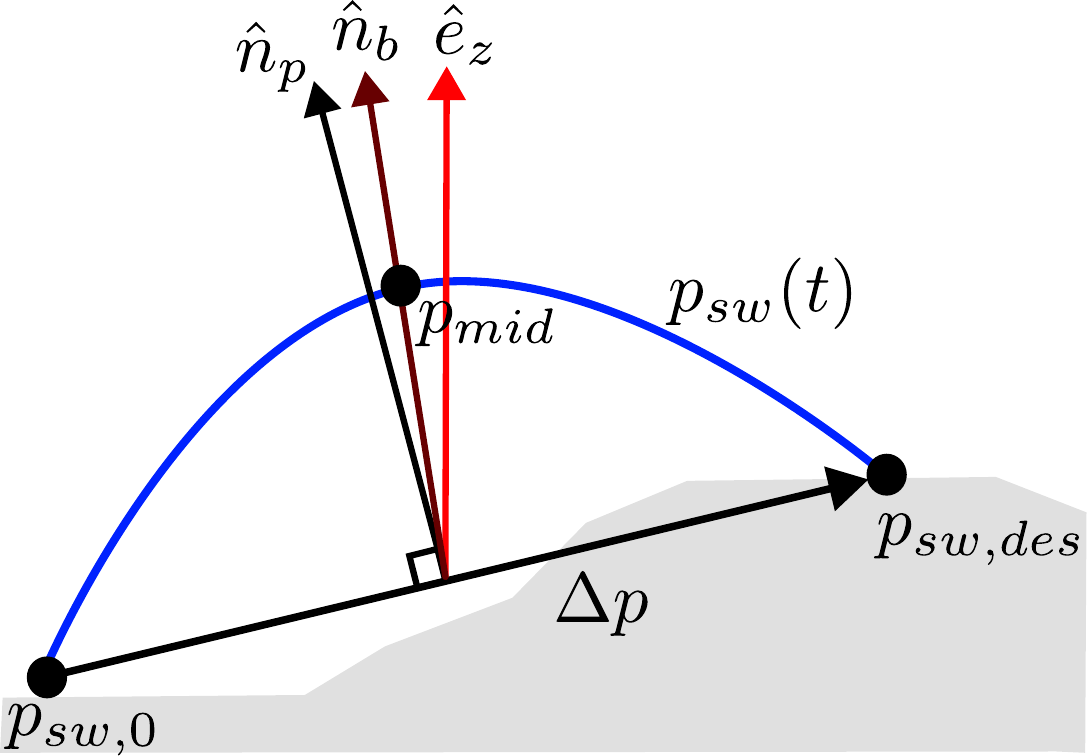}
	\caption{Trajectory from the swing foot position at the beginning of the swing phase, $p_{sw, 0}$ to the next footstep solution from MPFC, $p_{sw, des}$. We adapt the direction and clearance of the trajectory's midpoint, $p_{mid}$, based on the relative positions of $p_{sw, 0}$ and $p_{sw, des}$ to ensure sufficient ground clearance.}
	\label{fig:swing_traj}
\end{figure}

Our clearance scheme (\cref{fig:swing_traj}) updates the midpoint of the swing-foot trajectory by adapting its direction and clearance to the total displacement of the swing foot. 
This gives sufficient clearance when stepping up over steps without unnecessarily high steps on flat ground.

Let the swing foot position at the beginning of the swing phase be $p_{sw, 0}$,  the target foot position for the end of swing be $p_{sw, des}$, and define $\Delta p = p_{sw, des} - p_{sw, 0}$. 
We construct a unit vector $\hat{n}_{p}$ which is perpendicular to $\Delta p$ and lies in the plane spanned by $\Delta p$ and the world $z$ axis. 
When $\Delta p$ is small, for example when the robot is stepping in place, small variations in height estimates can lead to $\hat{n}_{p}$ pointing in inconsistent directions, therefore we blend $\hat{n}_p$ with the unit $z$-vector, $\hat{e}_{z}$ to get a blended direction, $\hat{n}_{b}$:

\begin{equation*}
		\hat{n}_{b} = (1 - s) \hat{e}_{z} + s \hat{n}_{p}\\
\end{equation*} 
where
\begin{equation*}
	s = \text{clamp}\left(\frac{\lVert \Delta p \rVert - 0.1}{0.1}, 0, 1\right).
\end{equation*}

The final waypoint location is then defined as 
$$p_{mid} = p_{sw, 0} + \frac{1}{2} \Delta p + c_{clear}\frac{\hat{n}_{b}}{\lVert \hat{n}_{b
	} \rVert}$$

where $c_{clear} = c + \min(c, \Delta p_{z})$ is the final swing foot clearance, and $c$ is a tuneable parameter representing the swing foot clearance on flat ground, which we set to 15 cm in our experiments. 

\begin{figure*}[btp]
	\centering
	\includegraphics[width=\textwidth]{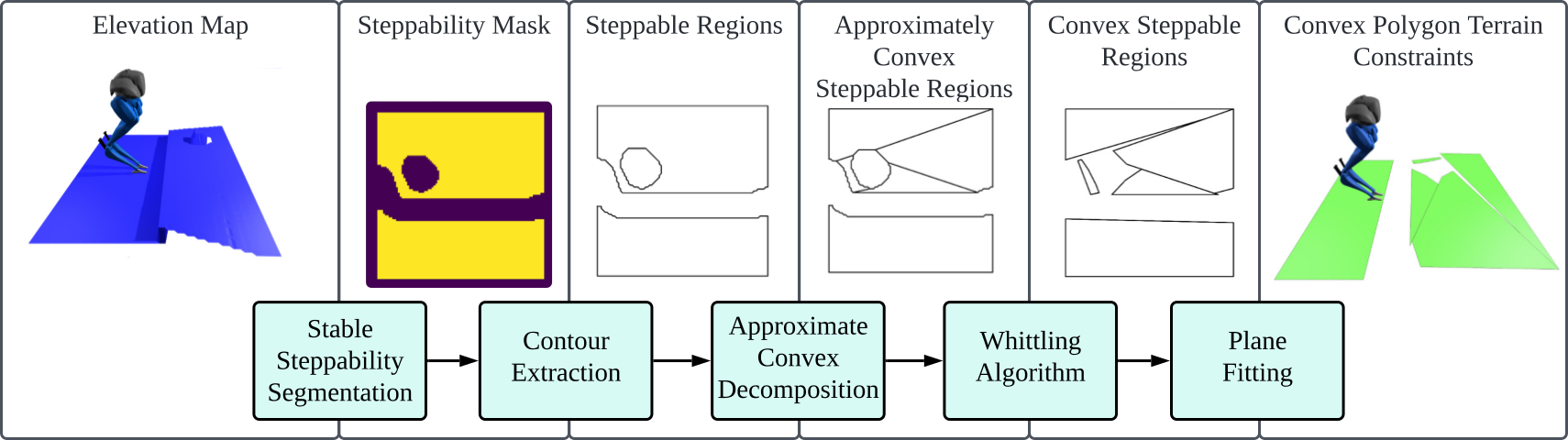}
	\caption{Pipeline for converting an elevation map of the terrain into a set of convex polygons for planning safe footsteps.
		Stable Steppability Segmentation produces a temporally-consistent steppability mask representing a 2D overhead view of the safe terrain. 
		We then extract the \brian{2D} boundaries of the safe terrain as non-convex polygons, which we decompose into convex polygons using an algorithm based on approximate convex decomposition. 
		\brian{Finally, we fit plane parameters to the convex polygons using the height of their vertices on the elevation map.}
	}
	\label{fig:s3cond}
\end{figure*}
\subsubsection{Swing foot Planning QP}
After finding the desired mid-spline waypoint $p_{mid}$, we solve \eqref{eq:swing_traj_qp} to update the swing foot trajectory to the new footstep target $p_{sw, des}$ and swing phase duration $T$. 
In addition to passing through the desired midpoint and ending at the target location, we constrain the swing foot trajectory to be continuous up to acceleration with the previously planned swing foot trajectory:

\begin{align}
	\text{minimize } \int_{0}^{T}\ddot{p}_{sw}(t)^{2} dt \span \span \nonumber \\
	\text{subject to } & p_{sw, k}(t_{k-1}) = p_{sw, k-1}(t_{k-1}) & p_{sw}(T) = p_{sw, des} \nonumber \\
	 & \dot{p}_{sw, k}(t_{k-1}) = \dot{p}_{sw, k-1}(t_{k-1}) & \dot{p}_{sw}(T) = 0 \nonumber \\
	 & \ddot{p}_{sw, k}(t_{k-1}) = \ddot{p}_{sw, k-1}(t_{k-1}) & \ddot{p}_{sw}(T) = 0 \nonumber \\
	 & p_{sw}(T/2)  = p_{mid} &
	 \label{eq:swing_traj_qp}
\end{align}

where $k$ indexes each OSC control cycle. 
We transcribe \eqref{eq:swing_traj_qp} as a QP which optimizes over the coefficients of a polynomial representing the swing foot trajectory. 
By using the initial swing foot position as $p_{sw, 0}$ at the beginning of the swing phase, we ensure that the trajectory starts at the initial swing foot position without needing to explicitly enforce that constraint for every control cycle. 

\subsection{Constant References}\label{subsec:constant_references}
We track a constant pelvis roll and pitch of zero, and a constant swing-leg hip yaw (abduction) angle of zero.
We track a commanded pelvis yaw rate from the remote control, and a swing toe angle so that Cassie's foot makes an angle of $\arctan{k_{x}}$ with the ground.

\subsection{Ankle Torque}
We add a quadratic on the difference between MPFC and OSC ankle torque commands. 

%% file: chapters/perception.tex
\section{Stable Steppability Segmentation and Convex Decomposition}\label{sec:perception}
Our control framework for walking over convex polygons requires an effective pipeline for approximating the safe terrain as convex polygons online (\cref{fig:block_diagram}A).
This section introduces our solution, ``Stable Steppability Segmentation'' (S3) and a complementary convex decomposition procedure similar to that used in \cite{acostaBipedalWalkingConstrained2023} (\cref{fig:s3cond}). 

S3 uses local information to classify the safety of each pixel in an elevation map, yielding a binary steppability mask of the terrain. 
We then perform contour extraction on this mask, and a 2D convex decomposition on the resulting steppable regions. 
Finally, we fit plane parameters to the resulting convex polygons using the elevation map.

\brian{In contrast to plane segmentation, we \textit{do not} subdivide or reject any steppable region based on its estimated normal or its error with respect to a best fit plane. This approach prevents localized frame-to-frame variations from having an outsized effect on the final segmentation, because there are no subdivision boundaries which might vary between frames, and localized outliers cannot trigger a subdivision or rejection of an entire region.}
Because safety criteria are local, we \brian{further enhance} temporal consistency by simply adding hysteresis to the classification of each pixel. 
We also use additional metrics beyond gradient or roughness to determine steppability, as discussed in \cref{subsec:segmentation}.
The simplicity of our approach allows the entire pipeline from elevation mapping to publishing convex polygons to run in real time on a single CPU thread. 
The remainder of this section explains S3 and our accompanying convex decomposition procedure in detail. 

\subsection{Stable Steppability Segmentation} \label{subsec:segmentation}
The goal of steppability segmentation is to determine where on the elevation map is safe to step.
Because the segmentation determines the foothold constraints for MPFC, it is important that the segmentation algorithm is

\begin{itemize}
	\item Temporally consistent
	\item Computed in real time
	\item Appropriately conservative. 
\end{itemize}

\begin{figure}[h]
	\centering
	\includegraphics[width=0.9\linewidth]{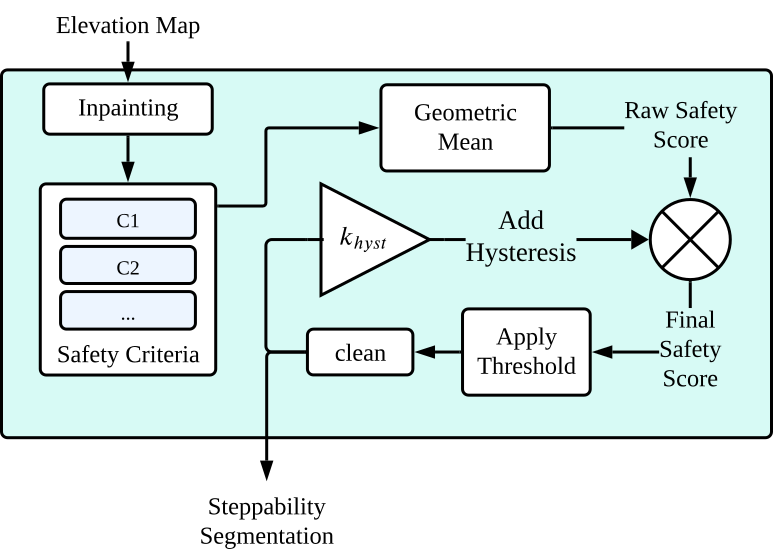}
	\caption{Block diagram of S3, our proposed terrain segmentation approach. Safety criteria are combined into an overall safety score for each elevation map pixel, before applying hysteresis to enhance temporal consistency. }
	\label{fig:s3}
\end{figure}

To accomplish this, we compute various ``safety criteria'' for whether a pixel is considered safe (\cref{fig:s3}). 
A safety criteria is a function transforming the elevation map to a pixel-wise ``safety score'' in the range $[0, 1]$, where 1 is completely safe, and 0 is unsafe. 
These safety criteria are fused via their geometric mean to yield an overall safety score. 
Temporal consistency is \brian{achieved through the local structure of the S3 algorithm,  as outlined above, and enhanced by} adding hysteresis to the overall safety score based on the previous segmentation. 
Realtime computation is achieved through the simplicity of our algorithm, and the small size of our elevation map. 
Because safety criteria are local to each pixel, S3 could also be GPU-parallelized for large elevation maps. 
The next subsection outlines what we mean by ``appropriately conservative'' and introduces a curvature-based safety criterion which accomplishes this goal. 
 
\subsubsection{Curvature Safety Criterion}
During our experiments in \cite{acostaBipedalWalkingConstrained2023}, we used a plane segmentation approach \cite{mikiElevationMappingLocomotion2022}, and had difficulty picking a safety margin which avoided tripping over curbs (\cref{fig:plane_margin}) while not taking excessively large steps over curbs. 
This experience illustrated the need to step further away from the bottom of a ledge than the top. 
To penalize terrain which is below edges, we need to identify terrain that is lower than its surroundings. 
Treating the elevation map as an image, this looks like applying a kernel that compares the height of each pixel to the average of the pixels around it:
\begin{figure}
	\centering
	\includegraphics[width=0.85\linewidth]{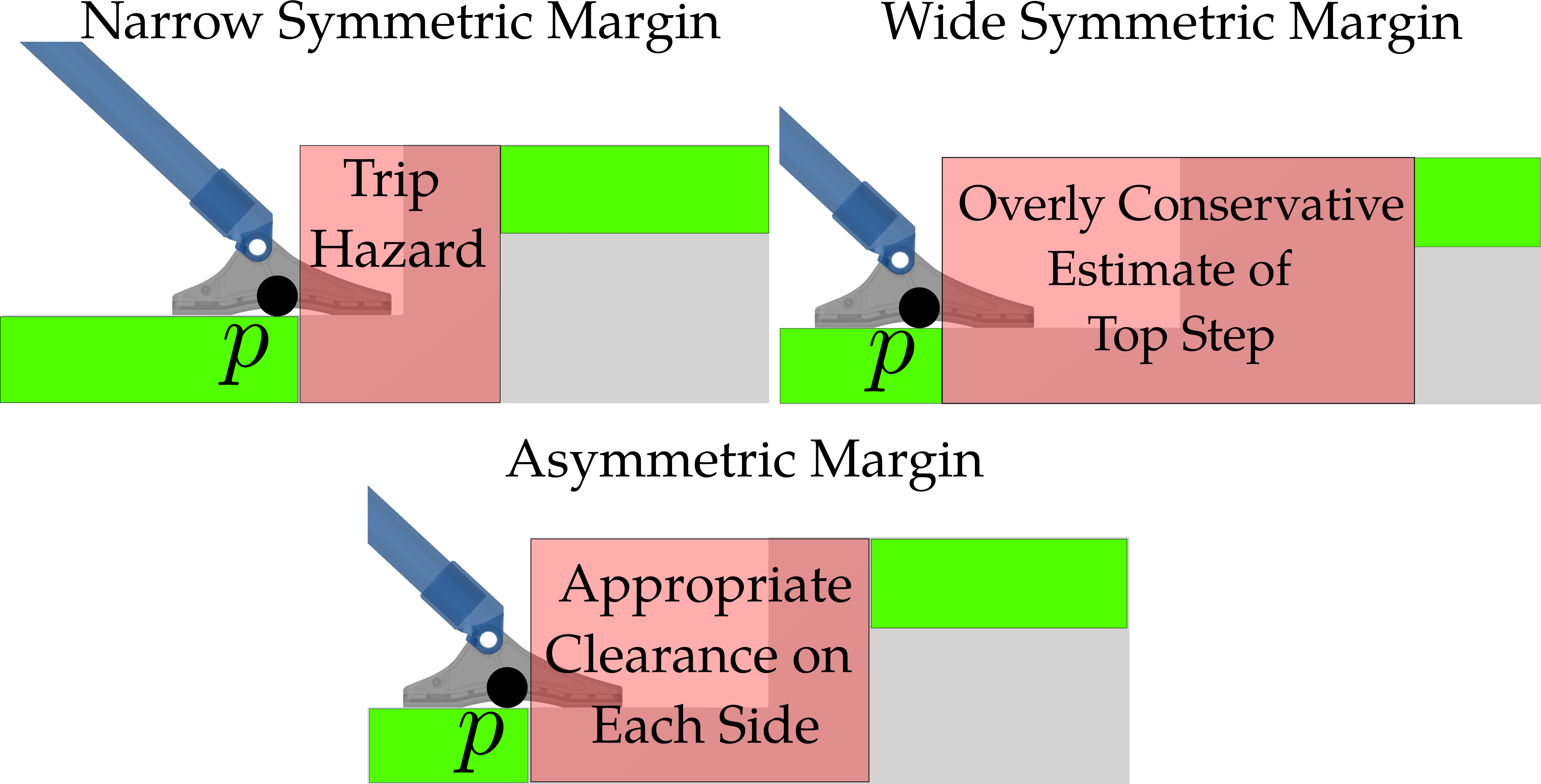}
	\caption{Walking over ledges with Cassie requires asymmetric constraints on the footstep position, $p$, which is mapped to the center of Cassie's foot. The desire to specifically avoid stepping below an edge motivated our curvature based safety criterion, \brian{which differentiates between sides of an edge using the sign of the elevation map's Laplacian.}}
	\label{fig:plane_margin}
\end{figure}
\begin{equation}
	\frac{1}{8}\begin{pmatrix}
		1 & 1 & 1 \\
		1 & -8 & 1 \\ 
		1 & 1 & 1
	\end{pmatrix}.
\end{equation}

This particular kernel is a Laplacian kernel, used to compute the curvature of an image, meaning we can use standard image processing tools to efficiently calculate this safety criterion.
Letting $E$ be the elevation map, the curvature criterion is computed via \cref{eq:curvature_criterion}, 
\begin{equation}
	c_{curve} = \min(1, \exp(-\alpha_{c}\text{LoG}(E)))
	\label{eq:curvature_criterion}
\end{equation}
where LoG is the Laplacian of Gaussian filter, which convolves the elevation map first with a Gaussian filter, then takes the Laplacian.
The pixel-wise exponential $\exp(-\alpha_{c} \text{LoG}(E))$ maps regions of positive curvature to the interval $(0, 1]$, with the score exponentially approaching $0$ as the curvature increases. 
The scale factor $\alpha_{c}$ tunes how aggressively positive curvature is punished. 
We take the min of the criterion with $1$ to ensure that no bonus points are awarded for negative curvature.    
\brian{Penalizing only positive curvature specifically targets area below edges, which poses a tripping hazard. 
To segment out the edge itself, we introduce an inclination safety criterion, which operates on the estimated normal of the elevation map to penalize steep terrain.}

\subsubsection{Inclination Safety Criterion}
The inclination safety criterion treats steep terrain as unsafe, by considering the magnitude of the $z$ component of the surface normal at each elevation mapping pixel. 
We estimate the normal using the covariance matrix of the positions around each pixel \cite{grandiaPerceptiveLocomotionNonlinear2022}, then square the  $z$ component to yield the inclination safety criterion,

\begin{equation}
	c_{inc} = n_{z}(E)^2.
\end{equation}  

To give context for how $c_{inc}$ classifies terrain in practice, we pick 0.7 as the final safety threshold for S3. 
For a pixel which is otherwise considered safe, this means a slope greater than about \ang{33} is unsafe, since $\cos^2(\ang{33}) = 0.7$.

\subsubsection{Combining Safety Criteria}
The final safety score is the geometric mean of the score for each criteria, plus a hysteresis value for all pixels classified safe in the previous frame. 
Pixels with a final score above some threshold are considered safe. 
The resulting binary image is post-processed to give a 2D view of the safe terrain around the robot. 
Letting $k$ index the time series of segmentations, the S3 output is given by
\begin{equation*}
	S_{k} =  \text{clean} \left( \left[ \left( \prod_{i = 1}^{M} c_{i}(E_{k}) \right)^{1/M} + k_{hyst}S_{k-1} \right] > k_{safe}\right)
\end{equation*}
where $M$ is the total number of safety criteria, and $\text{clean}(S) = \text{open}(\text{close}(\text{erode}(S)))$. 
\brian{The erode operation adds a safety margin to account for swing-foot tracking error and the length of Cassie's foot. 
The open and close operations remove any thin holes or protrusions. }

\subsection{Convex Planar Decomposition}\label{subsec:plane_decomposition}
\label{subsec:convex_decomp}
Finally, we convert the binary steppability mask into a set of convex planar polygons. 
We identify connected components of steppable terrain from the mask, and extract their outlines as 2D polygons.
In general, these are non-convex polygons with holes (caused, for example, by small obstacles or other unsteppable areas), but we require convex foothold constraints for the MPFC.
We use a two stage process to find a set of convex polygons whose union is an inner approximation these non-convex polygons.
This avoids creating many small triangles like an exact convex decomposition would, leading to fewer mixed integer constraints in the MPFC.

First, we perform approximate convex decomposition (ACD)\cite{lienApproximateConvexDecomposition2006} on each polygon.
ACD returns a decomposition of the original region into polygons which are $d$-approximately convex, where $d$ the depth of the largest concave feature.

After filtering out polygons with area less than 0.05 $\text{m}^{2}$, we find a convex inner-approximation of these nearly convex polygons with a greedy approach we name the whittling algorithm (\cref{alg:whittling}), after the way it makes incremental cuts to the polygon.
We initialize the output polygon, $\mathcal{P}$ as the convex hull of the original polygon, then take $\mathcal{P}$ to be the intersection of itself with greedily chosen half-spaces until no vertices of the original polygon are contained in the interior $\mathcal{P}$. To reduce the number of cuts we make, we initially sort the vertices by their distance to the boundary of $\mathcal{P}$, handling the innermost vertices first.

\begin{algorithm}[H]
	\caption{Whittling Algorithm} \label{alg:whittling}
	\begin{algorithmic}[0]
		\Require Input polygon vertices $V = \{v_{0}\ldots v_{n}\}$
		\Procedure{Whittle}{$V$}
		\State $\mathcal{P} \gets \text{ConvexHull}(V)$
		\State Sort $v_{i}$ by distance to $\partial\mathcal{P}$
		\ForAll{$v_{i}$}
		\If{$v_{i} \in \text{Interior}(\mathcal{P})$}
		\State $H$ = MakeCut($v_{i}$, $V$)
		\State $\mathcal{P} \gets \mathcal{P} \cap H$
		\EndIf
		\EndFor
		\State \Return $\mathcal{P}$ 
		\EndProcedure
	\end{algorithmic}
\end{algorithm}
\(\text{MakeCut}(V, v_{i})\) is a nonlinear program inspired by maximum margin classification \cite{boser1992training} which finds $a$ such that the half-space $H  = \{x \mid a^{T}(x - v_{i}) \leq 0\}$ contains as much of $V$ as possible:
\begin{align}
	\label{eq:make_cut}
	a = \underset{a}{\arg \min} & \sum_{j \neq i} \max(a^{T}(v_{i} - v_{j}), 0)^{2} \nonumber \\
	\text{subject to } & \lVert a \rVert_{2}^{2} = 1
\end{align}
We solve \eqref{eq:make_cut} using a custom gradient-based solver, which we detail in \cref{sec:whittling_solver}. 
Using the normal of the closest face of $\mathcal{P}$ to $v_{i}$ provides a high-quality initial guess for the solver.

To fit these polygons to the terrain, we project the 2D vertices onto the elevation map to recover the 3D position of each vertex. 
We then use least-squares to find the best fit plane to these vertices, yielding our final polygon representation. 

%% file: chapters/setup.tex
\section{Experimental Setup}\label{subsec:hardware}
This section explains the practical implementation of MPFC and our perception stack. 
The parameters used for S3, MPFC, and their supporting algorithms are given in \cref{sec:params_apendix}. 
The full perception and control system consists of six processes across three separate computers. 
Cassie's target PC runs a Simulink Real-Time application which publishes joint positions and velocities and IMU data, at 2kHz, and subscribes to torque commands. 
Communication between the target PC and Cassie's onboard Intel NUC occurs over UDP. 
The NUC runs the state estimator and a torque publisher to communicate with the target PC, and the OSC process, which includes the CoM and swing foot planner. 

The perception stack and MPFC are run on an off-board ThinkPad p15 Laptop with an 8-core, 2.3 GHz Intel 1180H processor and 24 GB of RAM.
The perception stack performs elevation mapping, terrain segmentation, and convex decomposition in one thread, and has a second thread to poll the Intel RealSense. 
The state estimator, operational space controller, torque publisher, perception stack, and MPFC communicate over LCM \cite{huangLCMLightweightCommunications2010} for low latency.

Except for the low-level target PC, all processes use the Drake systems framework \cite{russtedrakeandthedrakedevelopmentteamDrakeModelBasedDesign2019} to drive their operation. 
We solve the MPFC problem using Gurobi, and the OSC QP using FCCQP \cite{fccqp}.
\brian{We use the contact-aided invariant extended Kalman filter developed by Hartley et al. \cite{hartleyContactaidedInvariantExtended2020a} to estimate the pose and velocity of the floating base.}
Open source code for all of our contributed components will be provided in \href{https://github.com/DAIRLab/dairlib}{\texttt{dairlib}}\footnote{\url{https://github.com/DAIRLab/dairlib}}. 

\subsection{RealSense D455 Depth Camera}
The RealSense is mounted to Cassie's pelvis, pointed downward toward the terrain in front of the robot. 
We use \texttt{librealsense2} to subscribe to RealSense frames via a dedicated polling thread, with the perception stack thread accessing these frames through a shared buffer. 
We apply a decimation filter to reduce point cloud density. 

\subsection{Robot-Centric Elevation Mapping}\label{subsec:elevation_mapping}
We use the framework of Fankhauser et al.~\cite{fankhauserProbabilisticTerrainMapping2018} to construct a robot-centric elevation map of the terrain.
This framework represents the terrain as a regular grid, with the height of each cell updated by point cloud measurements through a Kalman filter.
Because Cassie's legs are visible in the camera frame, we crop out any points inside bounding boxes around Cassie's leg links.
\brian{State estimate z-drift is a well-known source of elevation mapping artifacts which must be corrected for an accurate terrain estimate.  
Related works use perception information to correct drift \cite{mikiElevationMappingLocomotion2022}, however this correction is not always sufficient when the walking motion generates non-negligible impacts \cite{grandiaPerceptiveLocomotionNonlinear2022}, as is the case for most Cassie walking controllers.}
To strongly correct for state estimate z-drift, before each point cloud update, we adjust the height of the elevation map by adding the height difference between the elevation map and the current stance foot.
\brian{To account for outliers, we calculate the elevation map height as the median of a 4x4 pixel grid, centered at the contact point.}

%% file: chapters/results.tex
\section{Results}\label{sec:results}
The perception and control architecture presented in this paper enables Cassie to walk over previously unseen terrain by identifying safe terrain and planning stabilizing footsteps subject to non-convex terrain constraints in real time.
This section presents experiments to show these capabilities and support our key claims. 
\brian{We perform simulation experiments to validate control design decisions, and quantify the performance gap between walking over known vs. online-identified safe terrain.}
On hardware, we showcase underactuated walking over discontinuous terrain with the Cassie biped, reporting consistent sub-10-millisecond solve times for MPFC.
\brian{We use data from real-world tests on multiple surface types} to show the improved temporal consistency and faster run time of S3 compared to explicit plane segmentation. 
Finally we summarize the capabilities of MPFC and S3 as a complete system, highlighting the performance improvements as a result of the contributions in this paper.

\brian{\subsection{Simulation Experiments}
This subsection details simulation experiments on complex terrains.
Our Drake  simulation includes Cassie's leaf springs, reflected inertia, motor curves, joint limits, effort limits, and full collision geometry.
MPFC is commanded a velocity of $[v_{x}, v_{y}] = [0.375, 0]$ m/s, and OSC is commanded a yaw rate proportional to the heading error.  
We show the idealized capabilities of MPFC via a simulation with ground-truth state and terrain information.
Compared to \cite{acostaBipedalWalkingConstrained2023}, where the most challenging terrain shown was 50 cm deep stairs, we show Cassie walking over an 8m $\times$ 23 cm beam, and up stairs with a depth of 27 cm and a rise of 15 cm. 
We then show a perceptive simulation, which simulates the invariant EKF \cite{hartleyContactaidedInvariantExtended2020a} to provide state estimates to OSC, MPFC, and the elevation mapping system, and uses a simulated depth sensor as input to the perception pipeline.
The S3 steppable area is more conservative than ground truth, resulting in a traversable beam width of 35 cm and stair depth of 40 cm with perception. }

\brian{\subsubsection{Step-Timing Optimization}
We demonstrate the importance of step-timing optimization by measuring the success rate of walking across randomly generated stepping stones with and without step-timing optimization. 
We generate a $5 \times 3$ grid of stepping stones, where each stone has a random height, length, width, and position offset. 
The minimum length and width are controlled by the parameter $d_{min}$, and the centers are offset from a nominal spacing of $d_{min}+ 21$cm. 
The stepping stone dimensions are uniformly distrubuted with the bounds given in \cref{tab:stepping_stone_offsets}. 
An example stepping-stone terrain can be seen in \cref{fig:stepping_stone_gt_example}. 
We sweep $d_{min}$ from 35 cm to 70 cm and compute the success rate for traversing 50 random terrains at each size, which we report in \cref{fig:step_timing_results}.

\begin{figure}[h!]
	\centering
	\includegraphics[width=\linewidth]{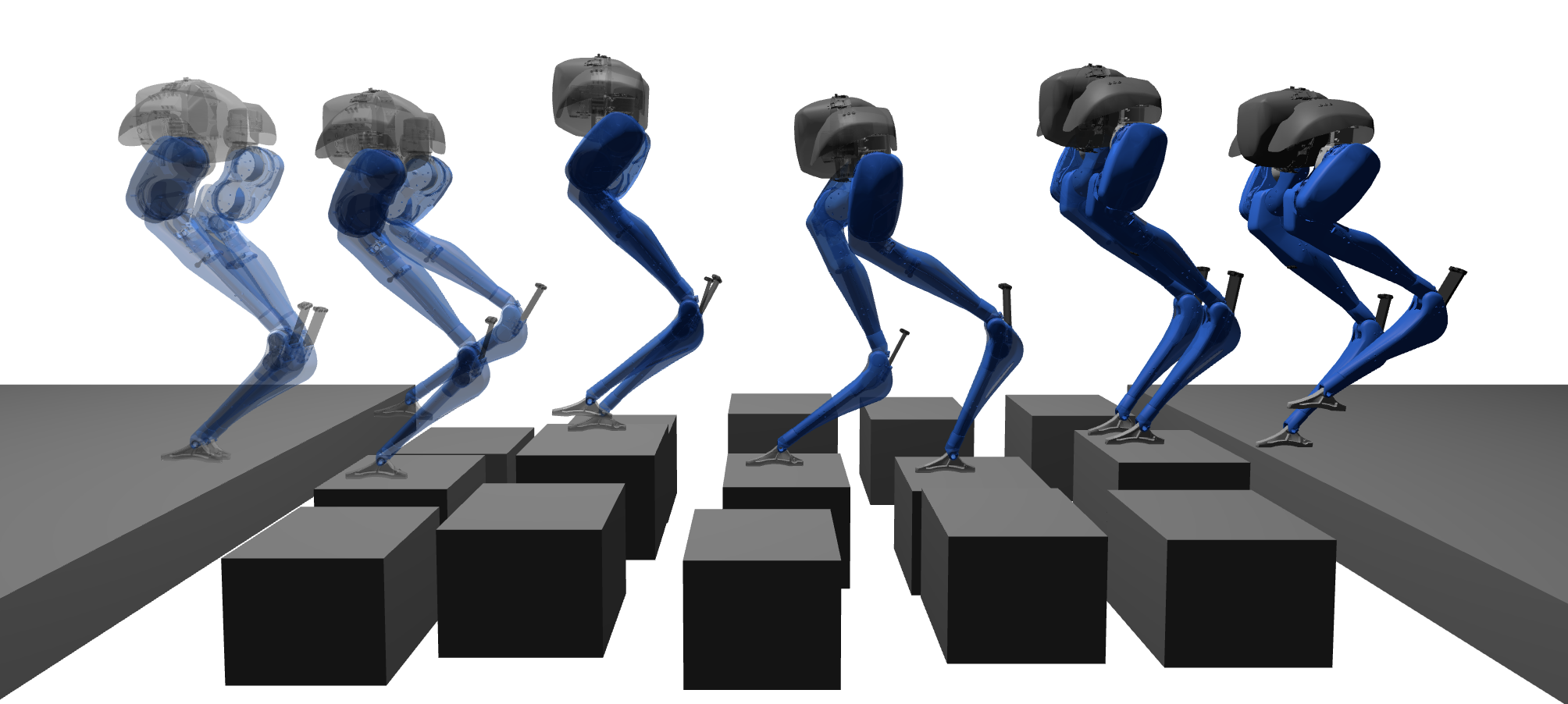}
	\caption{\brian{Example of successfully traversing a random stepping stone environment in simulation with $d_{min}$ = 35 cm.}}
	\label{fig:stepping_stone_gt_example}
\end{figure}

\begin{figure*}[t]
	\centering
	\includegraphics[width=\textwidth]{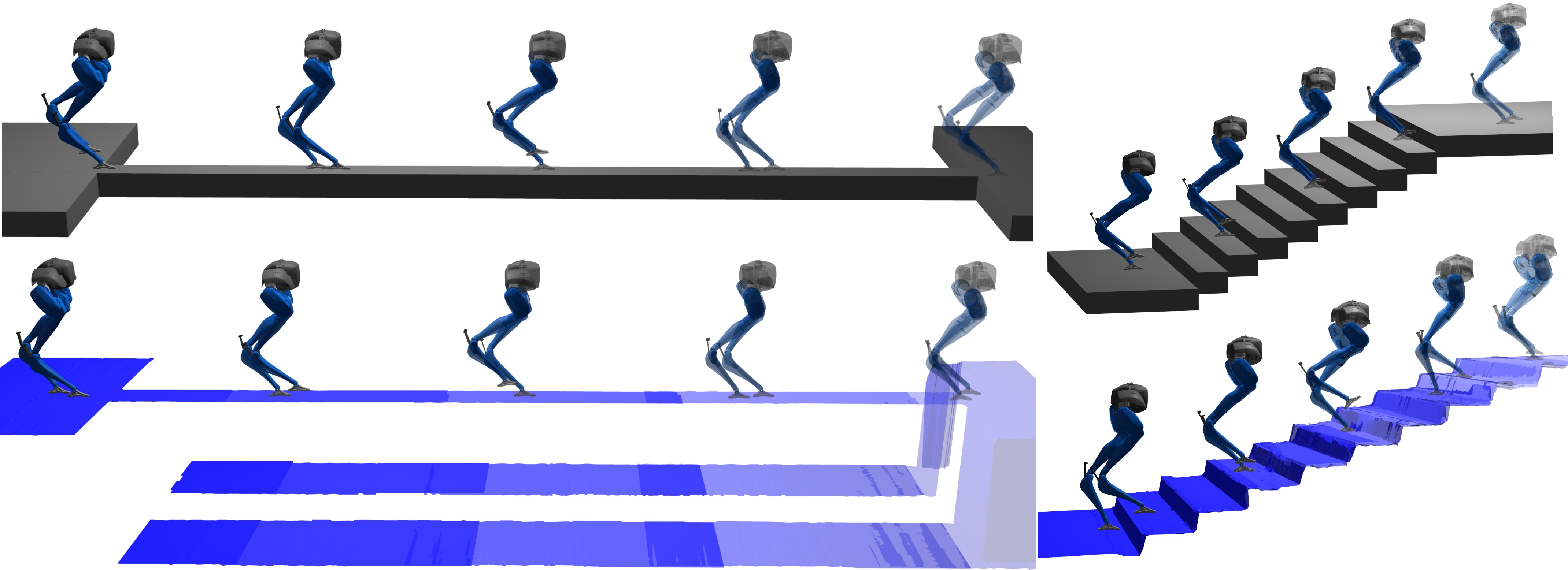}
	\caption{\brian{MPFC simulation experiments. \textbf{Top:} we use ground truth terrain information to walk over a 23 cm wide beam, and stairs with a rise of 15 cm and a depth of 27 cm. \textbf{Bottom:} Displaying the elevation map for walking over the same terrain types using S3.
			Safety margin in S3 results in less steppable area than for ground truth, so the minimum traversable dimensions are increased to 35 cm  wide for the beam and 40 cm deep for the stairs.}}
	\label{fig:sim_main}
\end{figure*}

\begin{figure}[h]
	\centering
	\includegraphics[width=0.9\linewidth]{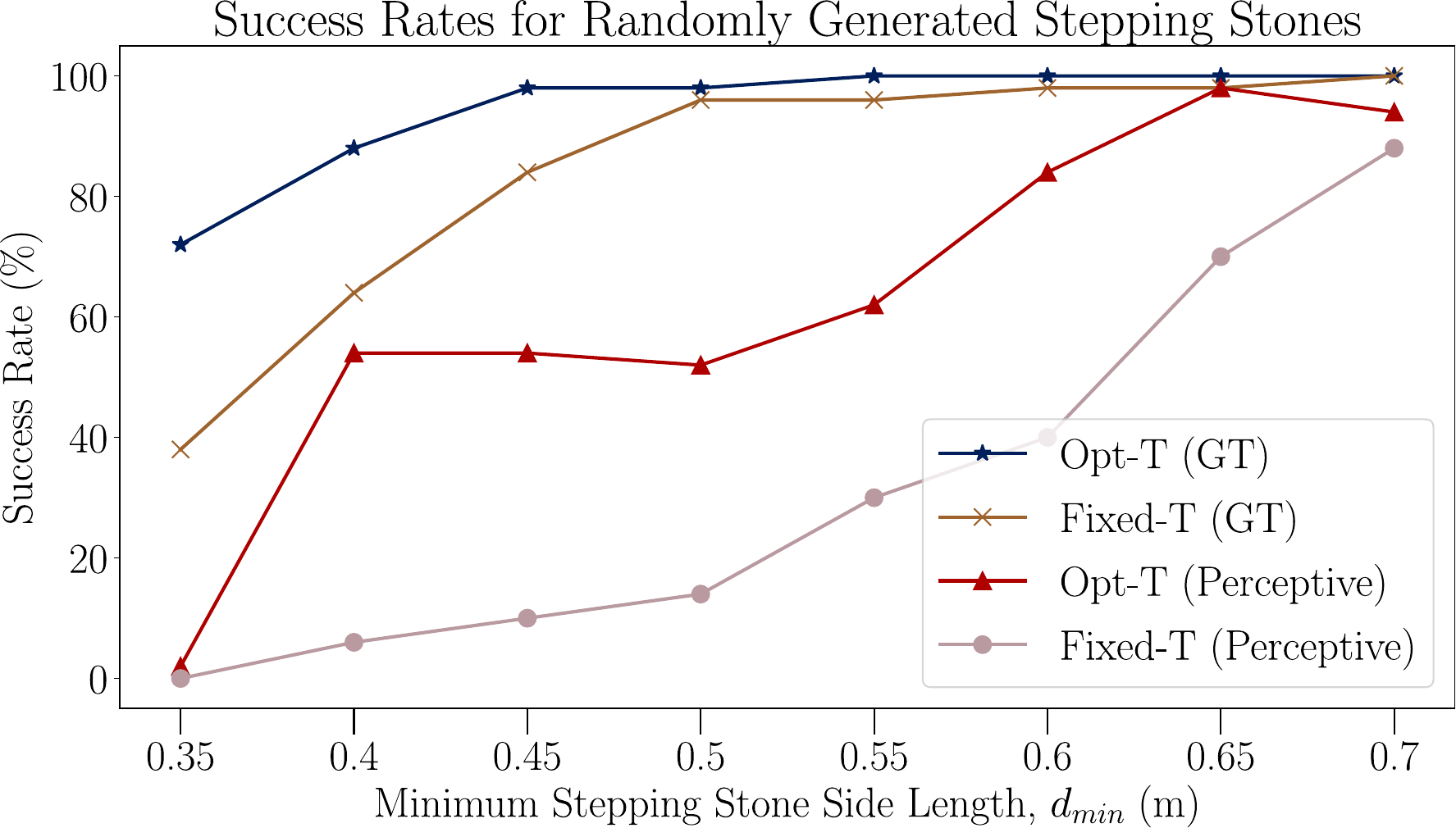}
	\caption{\brian{Success rates for walking across randomly-generated stepping stones in simulation. 
			Results with step-timing optimization are labeled Opt-T, and results without step timing optimization are labeled Fixed-T. 
			We report results with ground truth state and terrain (GT), as well as with perceptive terrain using S3 (Perceptive).
			Step-timing optimization increases success rates over small footholds. The lower success rate using perception is primarily due to isotropic safety margin in S3 reducing the lateral steppable area compared to ground-truth, which accounts for the width and length of the foot separately.}}
	\label{fig:step_timing_results}
\end{figure}

Step-timing optimization increases the success rate of walking over stepping stones, with larger effects for terrains with smaller footholds. 
Intuitively, underactuated dynamics strongly couple step-timing, stride length, and walking speed. 
Given Cassie's underactuation, step-timing therefore compensates for variability in the foothold location.
}

\brian{
\begin{table}[h]
	\centering
	\caption{Stepping Stone Parameter Distributions}
	\begin{tabular}{|l|c|}
		\hline
		\textbf{Parameter} & \textbf{Uniform Distribution Bounds}\\
		\hline
		$x$ offset & $\pm$ 5 cm \\
		\hline
		$y$ offset & $\pm$ 5 cm \\
		\hline
		$z$ offset  & $\pm$ 7.5 cm \\
		\hline
		length & $[d_{min}, d_{min} + 5]$ cm \\
		\hline
		width & $[d_{min}, d_{min} + 5]$ cm \\
		\hline
	\end{tabular}
	\label{tab:stepping_stone_offsets}
\end{table}
}

\subsection{Walking on Discontinuous Terrains}
\label{subsec:showing_terrains}
We show hardware experiments where Cassie walks over discontinuous and unstructured terrains using our perception and control stack. 
A single trial traversing steps, a curb, and a grass hill is shown in \cref{fig:results_hero}. 
Additional trials are shown in the supplemental video.

\subsection{Controller Solve Times}
\label{sec:solve_times}
To support our claims of faster than 100 Hz MPFC solve times, we compile solve times across 11:17 minutes of walking data from three experiments on the brick steps shown in \cref{fig:stairs-and-grass-motion-tiles}.
We give summary statistics of MPFC solve times in \cref{tab:mpfc_solve_time_stats}. 
This data uses a planning horizon of \brian{$N = 2$ footsteps, plus the initial single stance phase.}
The maximum solve time observed was 12.6 ms, with 99.9\% of solves taking less than 7.7 ms.
\brian{The median solve time was 2 ms, compared to the median solve time of of 5 ms and worst case solve time of over 20 ms reported in \cite{acostaBipedalWalkingConstrained2023} using the same computer.}
\input{results/solve_time_stats}
\blockcomment{
\begin{figure}[h!]
	\centering
	\includegraphics[width=0.9\linewidth]{svg-inkscape/solve_times_svg-tex}
	\caption{Solve-time distribution of MPFC.}
	\label{fig:mpfc_solve_times}
\end{figure}
}

\input{figures/results_hero}

\subsection{Perception Stack Evaluation}
\begin{figure}
	\centering
	\includegraphics[width=0.7\linewidth]{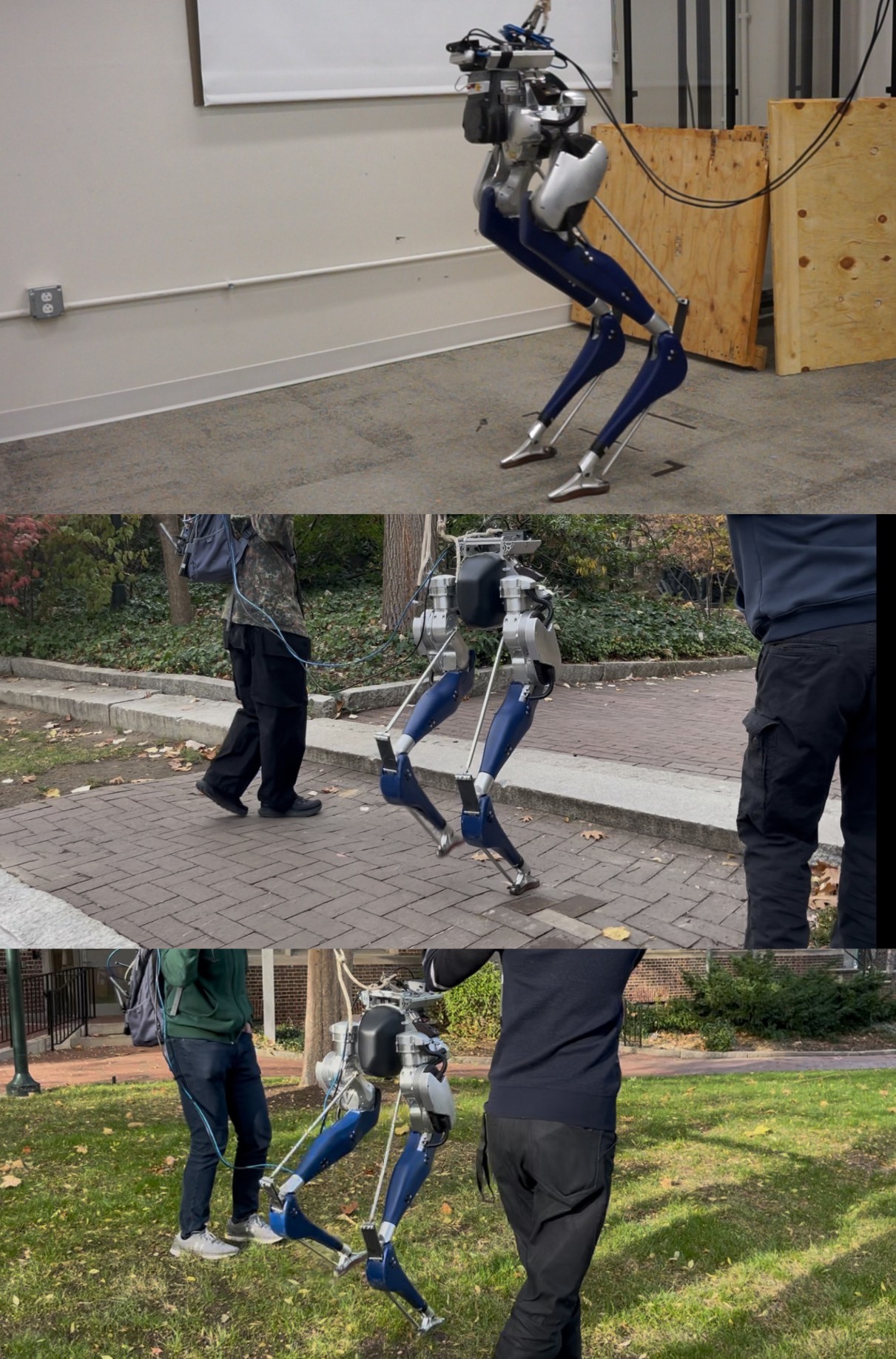}
	\caption{The environments used to collect data for benchmarking the perception stack's performance. 
		From top to bottom the terrains are \textbf{Lab}, \textbf{Brick Steps}, and \textbf{Grass}}
	\label{fig:perception_envs}
\end{figure}

This section supports our claims of S3's improved temporal consistency and faster run time compared to explicit plane segmentation. 
We use elevation mapping data from three terrains to evaluate segmentation performance (\cref{fig:perception_envs}). 
The \textbf{Lab} terrain establishes a baseline for each method in an ideal environment, where state estimate z-drift is the only potential challenge. 
The \textbf{Brick Steps} terrain features a set of brick steps where the bricks have settled over time, making the steps uneven, and unlikely to be segmented into a single plane by plane-segmentation methods. 
Similarly, the \textbf{Grass} terrain is challenging for plane segmentation approaches because our stance foot drift-correction conflicts with the height of the point cloud, introducing artifacts into the elevation map. 

We use the plane segmentation module developed by Miki at al. in the \texttt{elevation\_mapping\_cupy} software package \cite{mikiElevationMappingLocomotion2022} with default parameters (hereafter labeled EM\_cupy) as a plane segmentation baseline.
This algorithm has a similar structure to S3, starting by filtering the elevation map, classifying each cell as steppable or not, and then (unlike S3), trying to segment the steppable cells into planes. 
Each connected component of steppable terrain is checked for planarity, and if it fails, RANSAC \cite{Schnabel2007EfficientRF} is used to find smaller planes within that connected component. 
The authors of \cite{mikiElevationMappingLocomotion2022} also provide the option to disable RANSAC plane refinement, instead accepting or rejecting each connected component of steppable terrain in its entirety based on the estimated surface normal. 
We also test this variant, henceforth labeled EM\_cupy\_NR, where NR denotes ``No RANSAC'' or ``No Refinement.'' 
Because EM\_cupy\_NR is identical to the default EM\_cupy algorithm except for lacking a global planarity requirement on steppable regions, these results will support our argument that explicit plane segmentation is particularly brittle. 
\cref{tab:algo_comparison} summarizes the differences between S3 and the baselines. 

\begin{table*}[t]
	\centering
	\caption{Comparison of S3 and Plane Segmentation Baselines As Benchmarked}
	\begin{tabular}{|l||c|c|c|}
		\hline
		\textbf{} & \textbf{S3 (Ours)} & \textbf{EM\_cupy} & \textbf{EM\_cupy\_NR} \\
		\hline
		Steppability Criteria & Curvature, Inclination & Roughness, Inclination & Roughness, Inclination\\
		\hline
		Incorporates History & Yes & No  & No \\
		\hline
		Plane Refinement & None & RANSAC \cite{Schnabel2007EfficientRF} & Reject regions with slope $\geq$ \ang{30} \\
		\hline
		Inpainting Method & Navier-Stokes \cite{bertalmioInpaintNS} & Least Neighboring Value & Least Neighboring Value\\
		\hline
	\end{tabular}
	
	\label{tab:algo_comparison}
\end{table*}

\subsubsection{Computation Time}
\begin{figure}[t]
	\centering
	\includegraphics[width=0.95\linewidth]{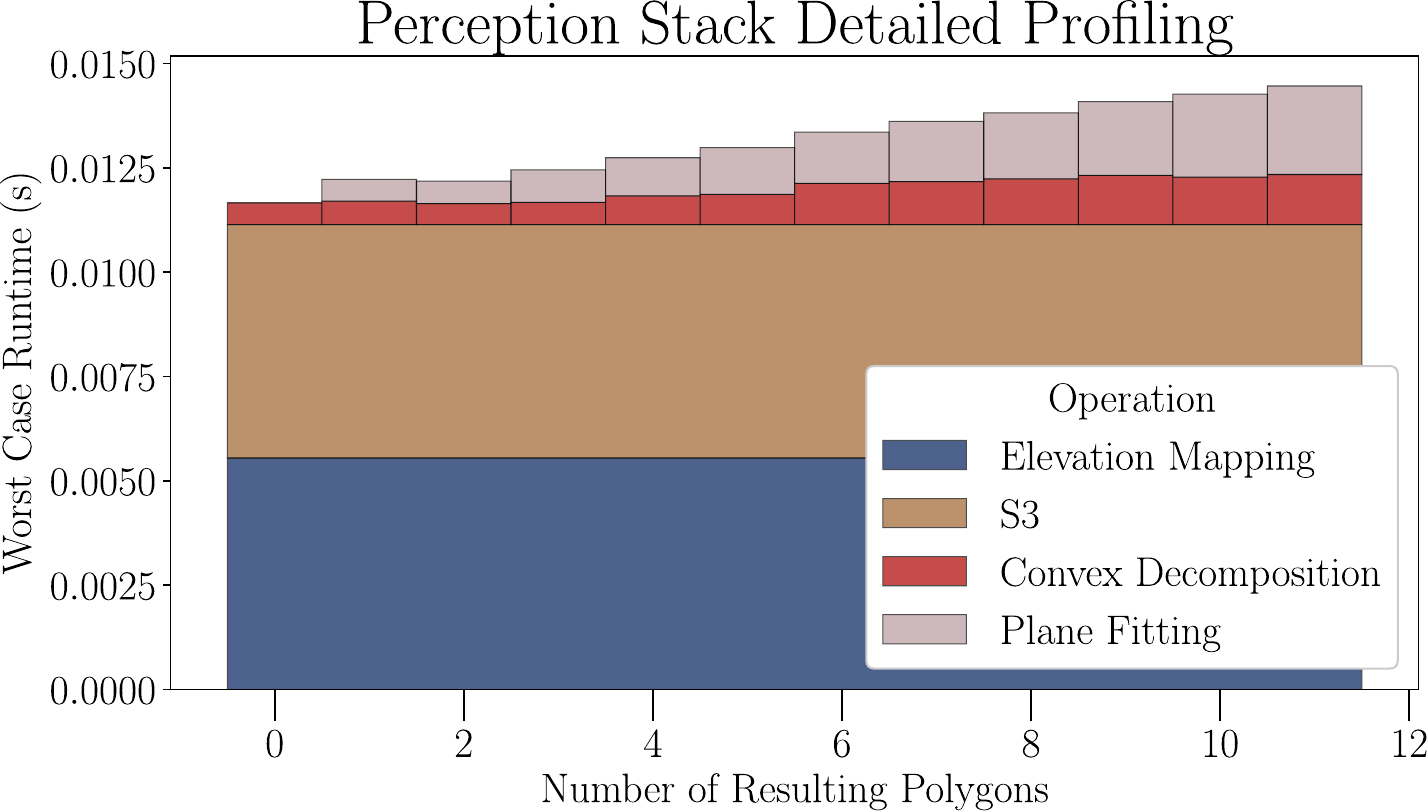}
	\caption{Detailed profiling of our perception stack, showing the worst-case runtime of each component. 
	``Convex Decomposition'' includes all steps necessary to convert the steppability mask from S3 into 2D convex polygons. 
	S3 and Plane Fitting use unoptimized python implementations, providing an avenue for further run time improvements. 
	Profiling is performed on the ThinkPad p15 laptop used for hardware experiments.}
	\label{fig:perception_profiling}
\end{figure}

We support the claim that our perception stack is real-time with detailed profiling.
To show that the entire pipeline is real-time, we profile the pipeline on 90 seconds of walking data from \textbf{Brick Steps}. We report the worst-case observed computation time for each step of the perception stack in \cref{fig:perception_profiling}, breaking the convex decomposition steps out by the number of resulting polygons. We note that the worst-case cumulative compute times stay below the 33~ms required for keeping up with the RealSense frame rate. 

As a benchmark against other segmentation approaches, we compare the run times of S3, EM\_cupy, and EM\_cupy\_NR for each test environment in \cref{fig:s3_vs_plane_results}, and find S3 to be the most consistent, with the lowest computation time.

\subsubsection{S3 Temporal Consistency}
We measure the temporal consistency of each segmentation approach via the intersection over union (IoU) of consecutive segmentations. 
IoU measures the ratio of pixels labeled as safe in both segmentation frames to the number of pixels labeled safe in either frame. 
Because data is lost when the elevation map moves relative to the world, we restrict the IoU computation to pixels which are present in both frames. 
A frame-to-frame IoU of 1 represents perfect temporal consistency, and 0 represents no overlapping safe terrain between segmentations.

\begin{figure*}[t!]
	\centering
	\includegraphics[width=0.98\linewidth]{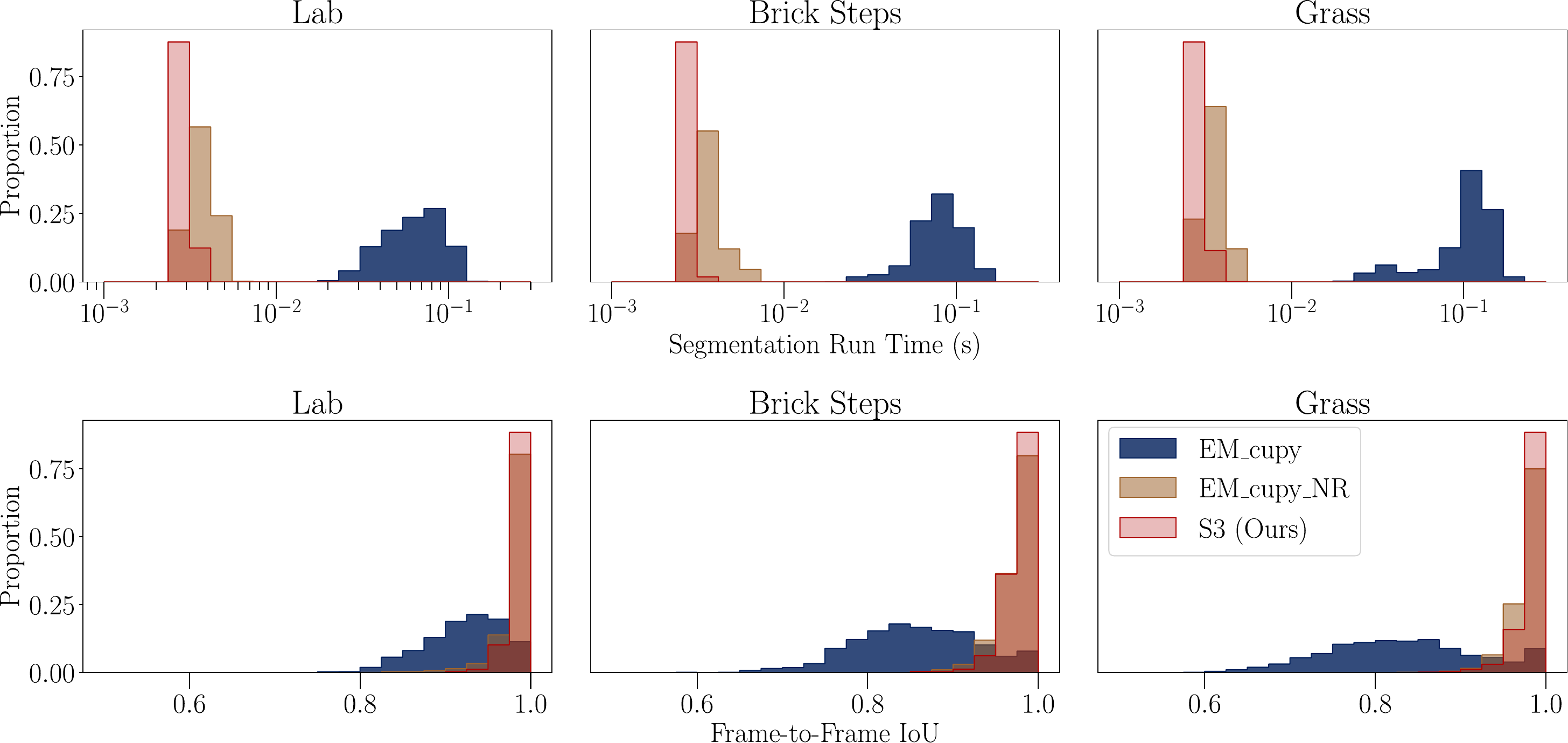}
	\caption{\briannew{Offline benchmark of S3 compared to plane segmentation baselines. }\textbf{Top:} Histogram of the run time of each segmentation algorithm over 60 seconds of elevation mapping data from each test environment. \briannew{Benchmark was run on an Apple Macbook Pro with an M1 Max CPU, 10 cores, and 64 gb of RAM. } S3 has the fastest and most consistent run times. EM\_cupy is the slowest, with a highly variable run time, due to the repeated use of RANSAC to refine the plane segmentation.
	\textbf{Bottom:} Histogram of the frame-to-frame IoU of the safe terrain segmentation over the same datasets. Our method reliably achieves a frame-to-frame IoU close to 1 across environments, representing excellent temporal consistency. }
	\label{fig:s3_vs_plane_results}
\end{figure*}

The distributions of frame-to-frame IoU for one minute of walking data in each environment are shown in \cref{fig:s3_vs_plane_results}.
Our approach consistently achieves an IoU close to 1 across environments, representing excellent temporal consistency, while EM\_cupy has a \briannew{notably lower IoU even in the lab setting, where elevation map artifacts from floating base drift result in som non-planarity in the map}. 

\briannew{EM\_cupy\_NR has similar temporal consistency to S3, though the lack of hysteresis contributes to small holes which appear and disappear.
The improved temporal consistency of EM\_cupy\_NR over the default EM\_cupy shows that the plane-segmentation step in particular is brittle, rather than other design choices like inpainting or steppability criteria.}
The segmentation output from each algorithm at 1 second intervals is shown in \cref{fig:seg_tiles}, and animations of the segmentation state are shown in the supplemental video.

\begin{figure*}[h!]
	\centering
	\includegraphics[width=0.3\textwidth]{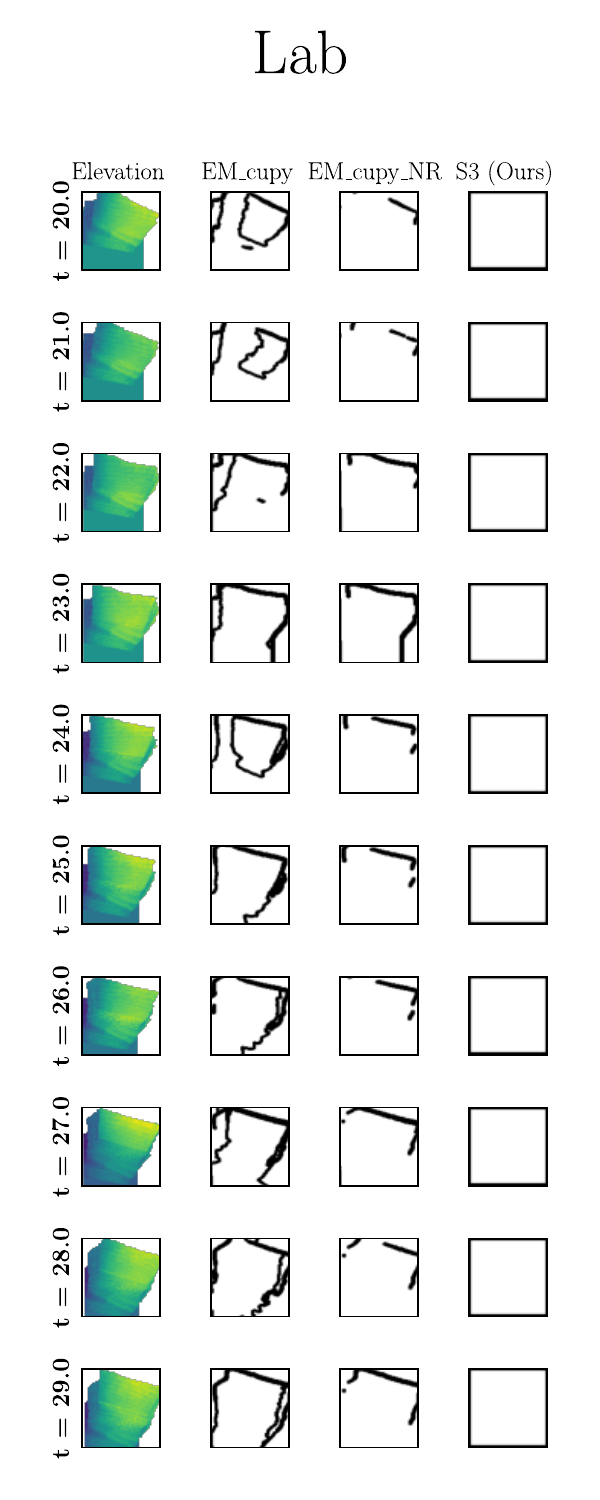}
	\includegraphics[width=0.3\textwidth]{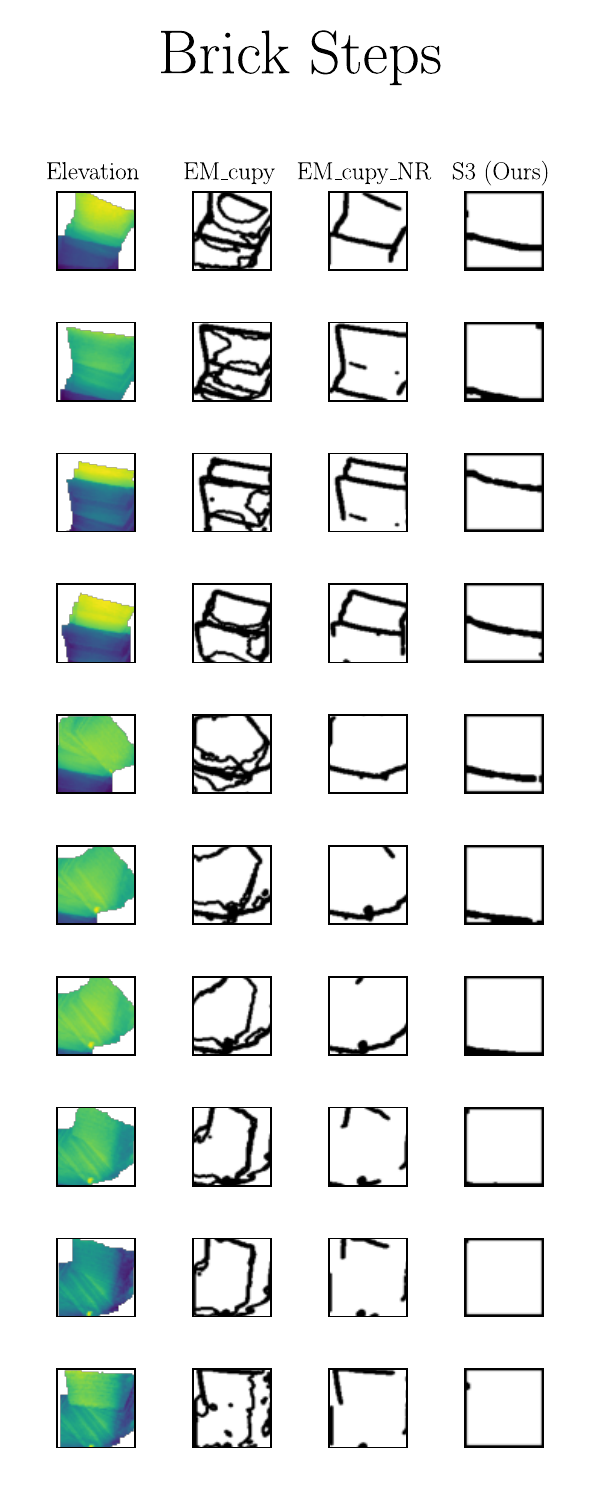}
	\includegraphics[width=0.3\textwidth]{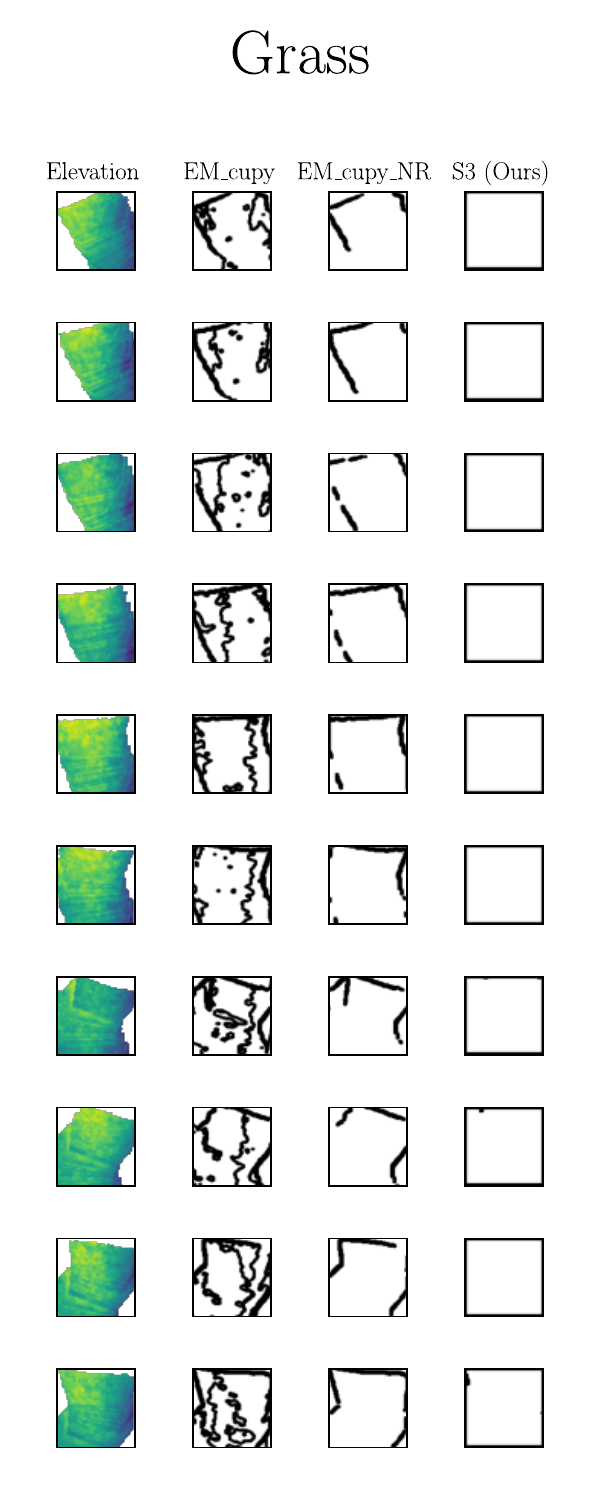}
	\caption{Tiles showing the output of each segmentation method for each evaluation environment at 1 second intervals. In the \textbf{Lab} and \textbf{Grass} environments, the use of Navier-Stokes based inpainting allows S3 to correctly identify the entire elevation map as steppable.  \briannew{The S3 segmentation experiences minimal ``flickering'' of the steppable terrain compared to the baselines}. Animations of these segmentation results can be seen in the supplemental video.}
	\label{fig:seg_tiles}
\end{figure*}

\subsubsection{\brian{Convex Polygon Temporal Consistency}}
\begin{figure}
	\centering
	\includegraphics[width=0.75\linewidth]{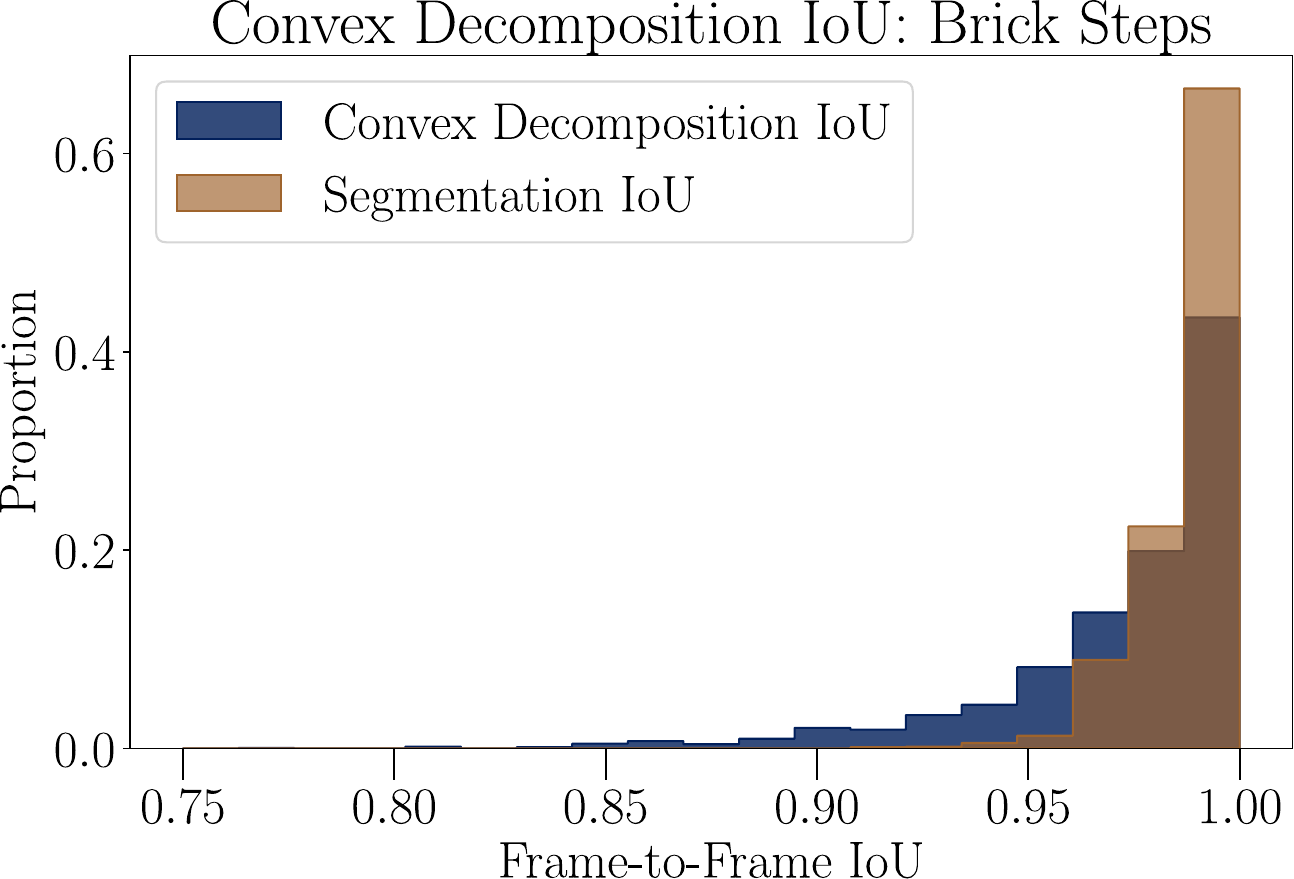}
	\caption{\brian{The final convex decomposition has a similarly shaped IoU distribution to the safe terrain segmentation, though is less consistent overall.}}
	\label{fig:decomposition_iou}
	\vspace{-0.5cm}
\end{figure}

\brian{
This section verifies that a temporally consistent terrain segmentation ultimately leads to a temporally consistent convex polygon decomposition. 
Because MPFC is free to pick any foothold for each solve, we compute the frame-to-frame IoU of the terrain covered by each convex decomposition, rather than the consistency of individual polygons. 
We compute the IoU by sampling. 
Each elevation map cell is marked as safe if its 2D position is covered by a convex polygon.
We then compute the IoU of the safe cells corresponding to each consecutive convex decomposition.
The distribution of IoU for the safe terrain vs. for the convex decomposition is shown in \cref{fig:decomposition_iou}.}

\brian{\begin{table}[h]
		\centering
		\caption{Moving Obstacles Segmented vs. Hysteresis}
		\begin{tabular}{|l|c|c|c|c|c|c|c|}
			\hline
			$k_{hyst}$ & 0.0 & 0.1 & 0.2 & 0.3 & 0.4 & 0.5 & 0.6 \\
			\hline
			Standing & 3/3 & 3/3 & 3/3 & 3/3 & 3/3 & 2/3 & 0/3 \\
			\hline
			Walking & 3/3 & 3/3 & 3/3 & 3/3 & 3/3 & 2/3 & 0/3 \\
			\hline
		\end{tabular}
		\label{tab:moving_obstacles}
	\end{table}}

\subsubsection{S3 Hysteresis and Moving Obstacles}
\begin{figure}[h]
	\centering
	\includegraphics[width=0.8\linewidth]{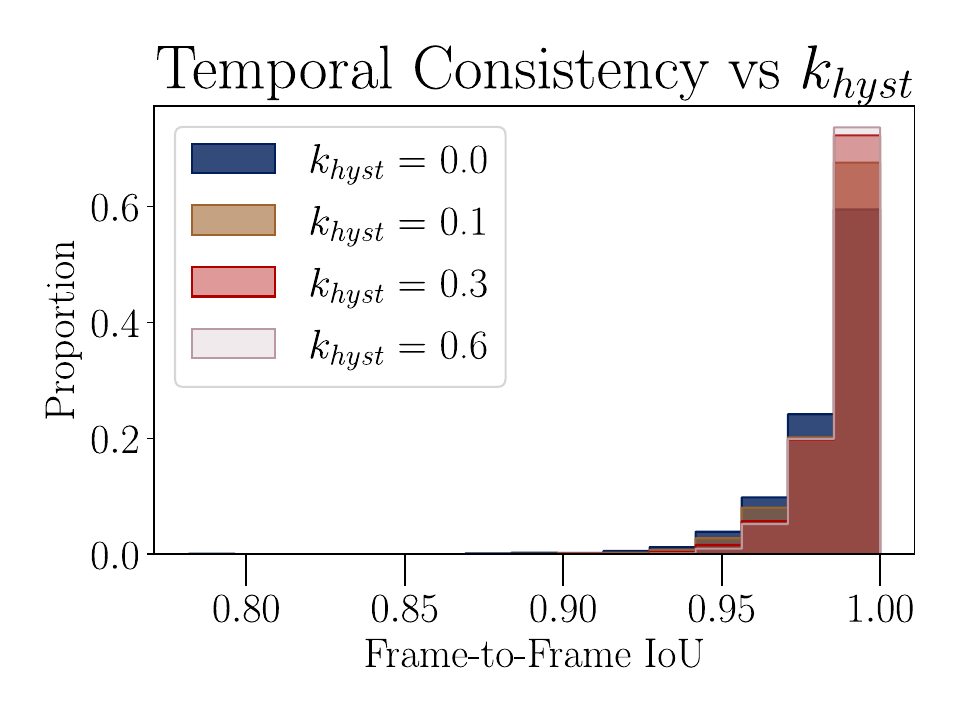}
	\caption{Frame-to-Frame IoU of the S3 terrain segmentation results with varying levels of hysteresis, evaluated on the \textbf{Brick Steps} data. \brian{The lowest observed IoU was 0.78, even without hysteresis, in contrast to both of the plane segmentation baselines, whose the IoU varied across the entire [0, 1] interval.} }
	\label{fig:hist_iou}
\end{figure}

For our hardware experiments, we chose a hysteresis value of 0.6, with a safety threshold of 0.7.  
This high level of hysteresis enhanced the \brian{consistency} of the segmentation for the terrains we tested on, however less hysteresis may be desirable in dynamic environments. 
\brian{To determine the appropriate hysteresis for different scenarios, we provide two analyses. 
	First, we perform experiments with moving obstacles, by tossing 15 cm foam cubes into the scene and counting how many are segmented out for each hysteresis condition. 
	An obstacle is defined as segmented out if it creates a hole in the terrain segmentation before it comes to rest.
	These results are shown in \cref{tab:moving_obstacles} and in the supplemental video.
	Second, we show the distribution of frame-to-frame IoU values for varying levels of hysteresis on the \textbf{Brick Steps} in \cref{fig:hist_iou}. 
	The lowest observed IoU was 0.78, even without hysteresis, and a hysteresis factor as low as 0.3 performed similarly to the chosen value of 0.6. 
	These results suggest that a hysteresis factor between 0.3 and 0.4 can enhance the stability of the terrain segmentation while maintaining correctness in dynamic environments.}

\subsection{Summary of Capabilities}
We briefly summarize the capabilities of our perception and control architecture, and the performance improvements from MPFC and S3.
Compared to our deadbeat ALIP footstep planner based on \cite{gongOneStepAheadPrediction2021}, MPFC is more robust, even on hard, flat surfaces, due to the inclusion of workspace constraints, ankle torque, and step timing optimization.
MPFC and S3 also enable Cassie to walk up and down steps and curbs up to 16 cm tall when each step is deep enough to have a valid S3 segmentation. 
In contrast to our original perception implementation in \cite{acostaBipedalWalkingConstrained2023}, using S3 for terrain segmentation allows the robot to walk continuously with perception in the loop due to its temporal consistency, \brian{despite} artifacts from state-estimate drift and impacts.
This is the case even on grass, where proprioceptive and exteroceptive ground height estimates conflict. 
We discuss limitations of our stack in \cref{subsec:limits}.

%% file: results/solve_time_stats.tex
\begin{table}[h!]
\centering
\caption{MPFC Solve-Time Statistics (134,654 Solves)}
\begin{tabular}{|c|c|c|c|}
\hline
Mean & Median & 99.9th Percentile & Maximum\\
\hline
0.0022 & 0.0020 & 0.0077 & 0.0126 \\
\hline
\end{tabular}
\label{tab:mpfc_solve_time_stats}
\end{table}

%% file: figures/results_hero.tex
\begin{figure*}
	\centering
	\begin{subfigure}[t]{\textwidth}
		\centering
		\includegraphics[width=0.115\textwidth]{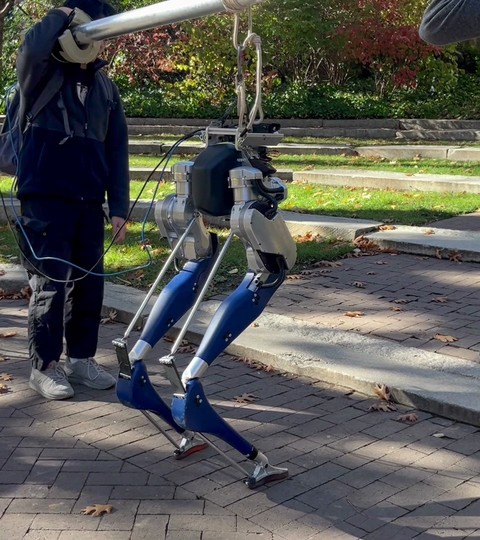}
		\includegraphics[width=0.115\textwidth]{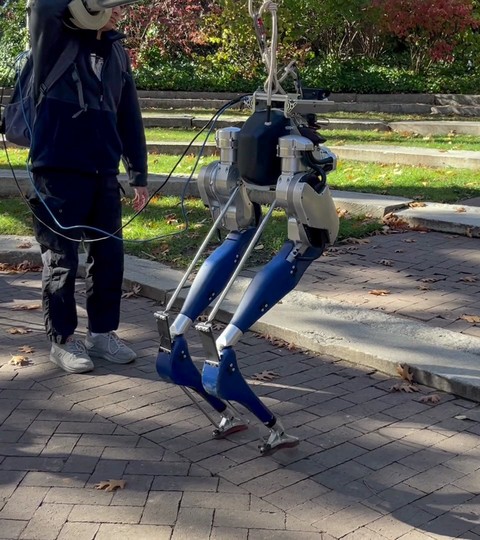}
		\includegraphics[width=0.115\textwidth]{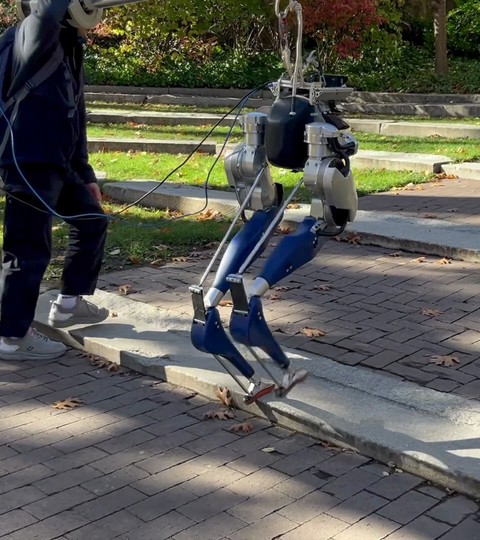}
		\includegraphics[width=0.115\textwidth]{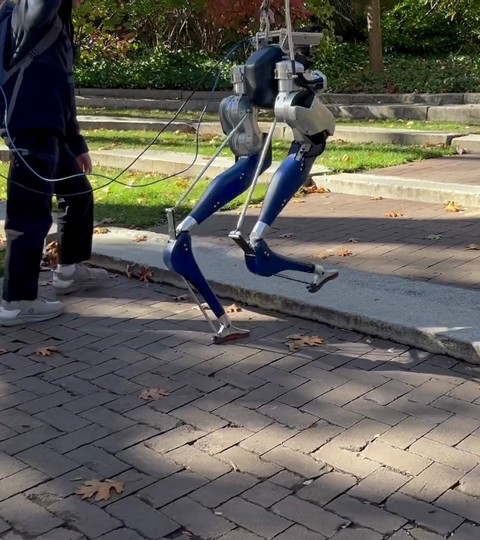}
		\includegraphics[width=0.115\textwidth]{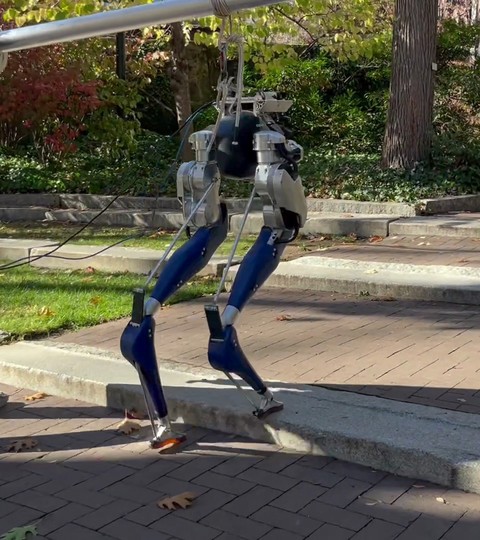}
		\includegraphics[width=0.115\textwidth]{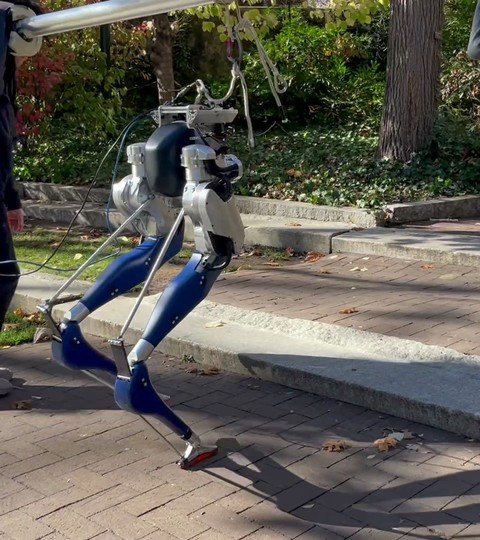}
		\includegraphics[width=0.115\textwidth]{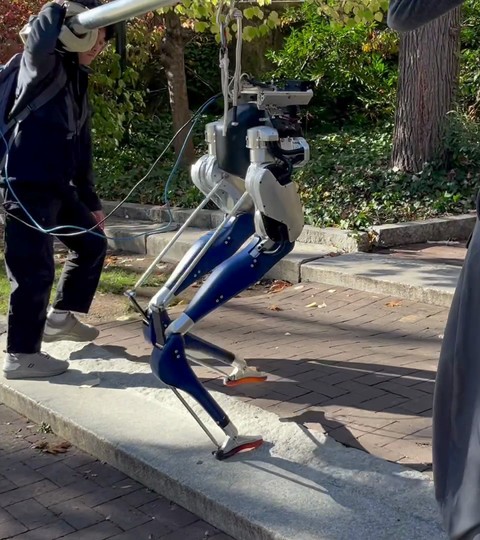}
		\includegraphics[width=0.115\textwidth]{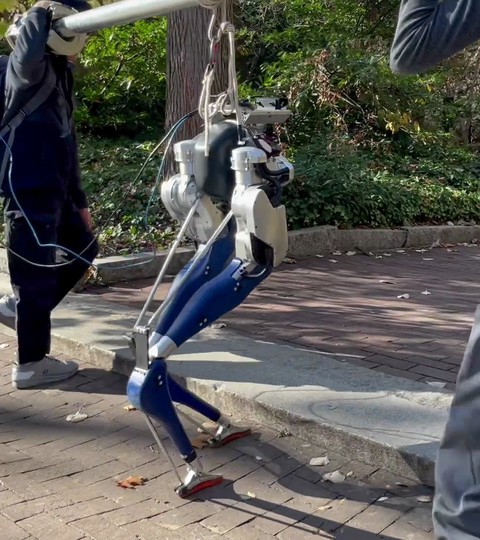}\\
		\vspace{3pt}
		\includegraphics[width=0.115\textwidth]{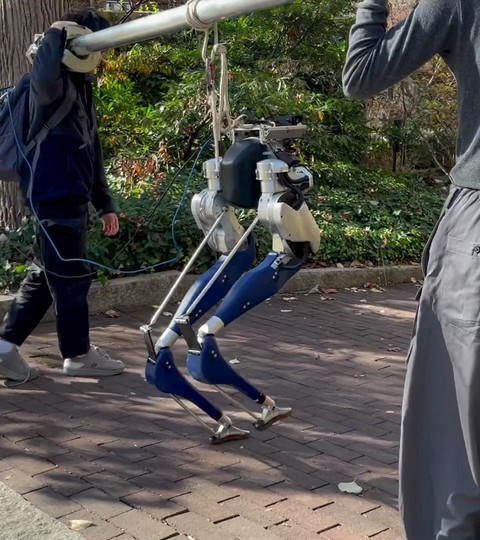}
		\includegraphics[width=0.115\textwidth]{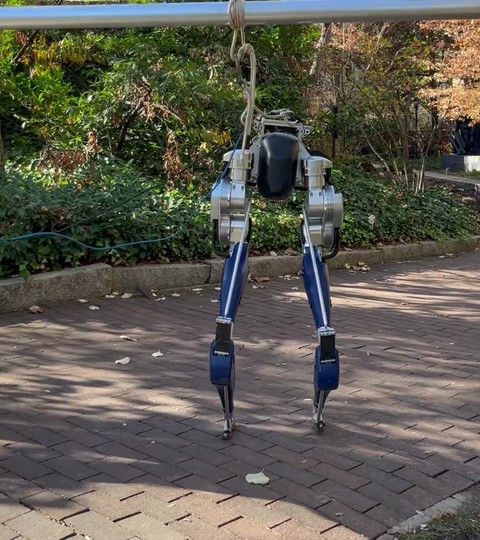}
		\includegraphics[width=0.115\textwidth]{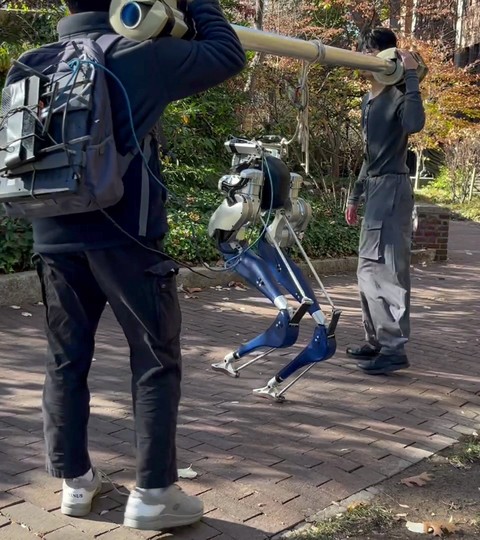}
		\includegraphics[width=0.115\textwidth]{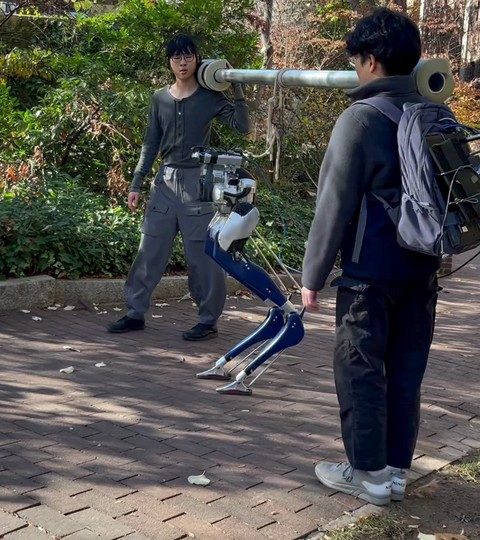}
		\includegraphics[width=0.115\textwidth]{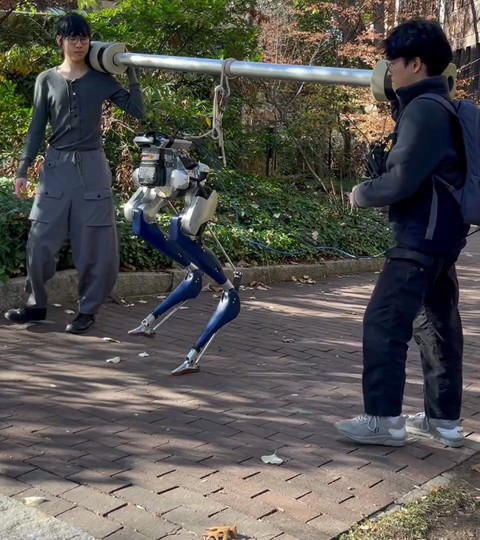}
		\includegraphics[width=0.115\textwidth]{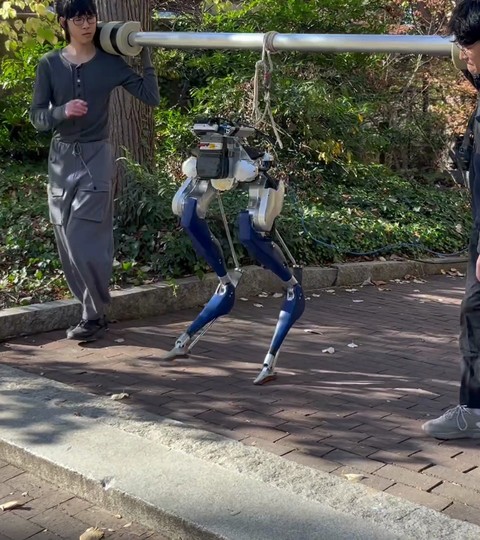}
		\includegraphics[width=0.115\textwidth]{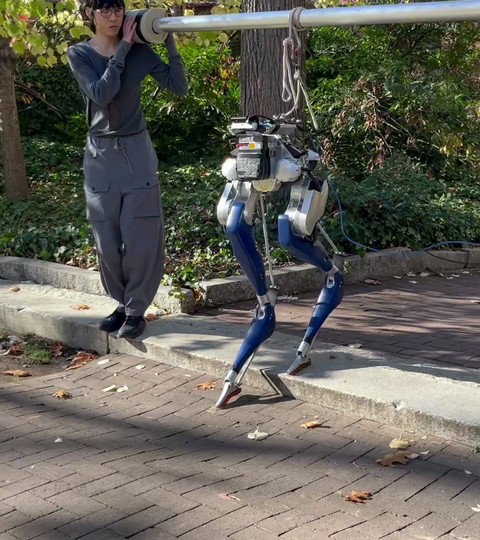}
		\includegraphics[width=0.115\textwidth]{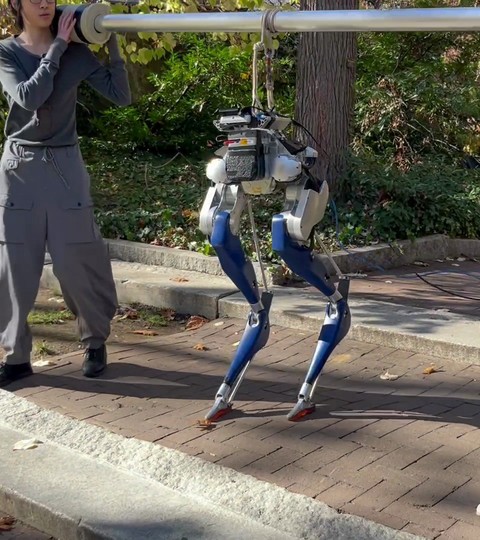}\\
		\vspace{3pt}
		\includegraphics[width=0.115\textwidth]{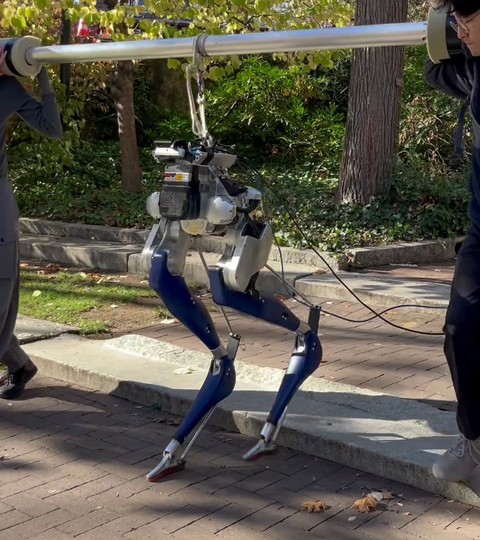}
		\includegraphics[width=0.115\textwidth]{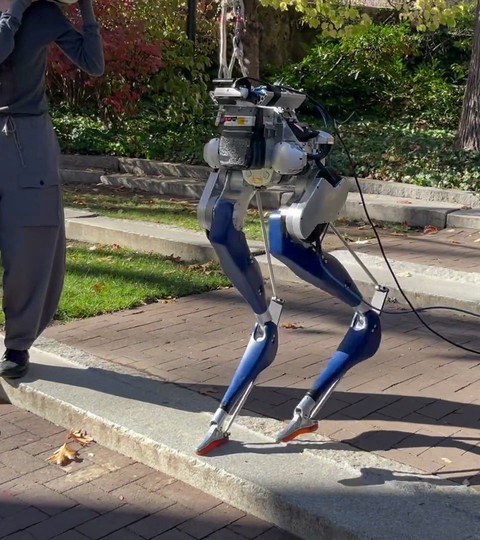}
		\includegraphics[width=0.115\textwidth]{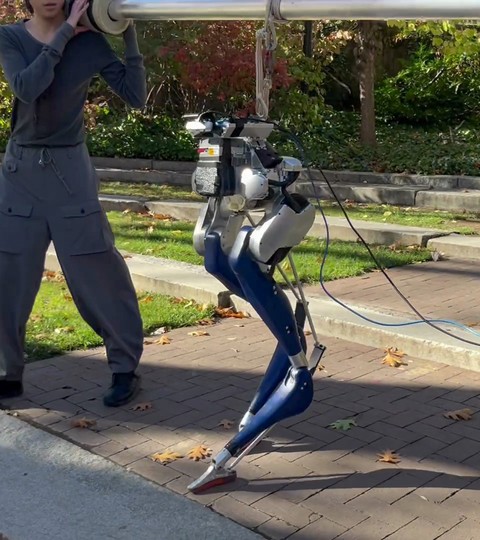}
		\includegraphics[width=0.115\textwidth]{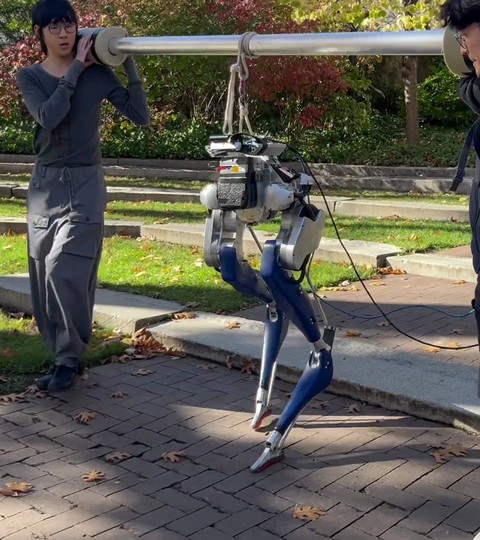}
		\includegraphics[width=0.115\textwidth]{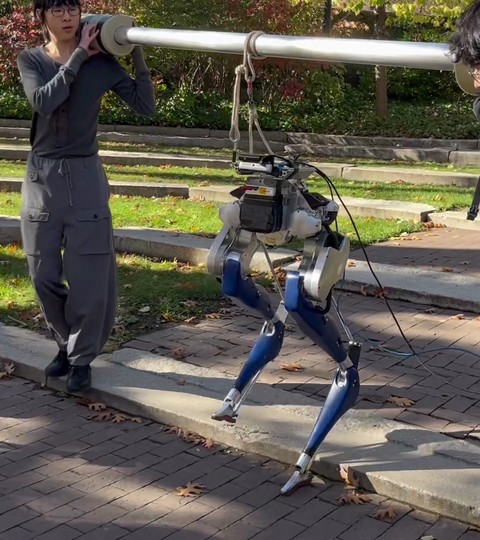}
		\includegraphics[width=0.115\textwidth]{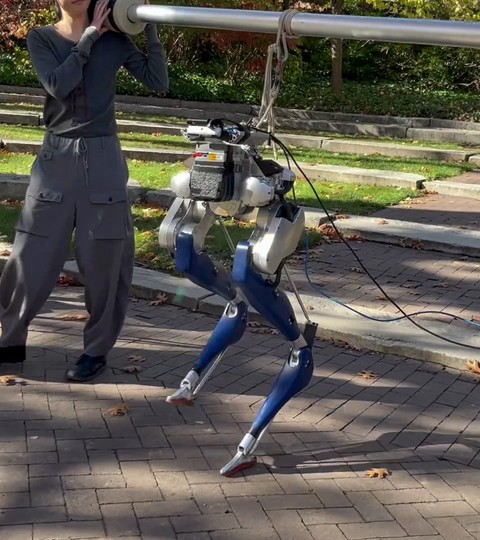}
		\includegraphics[width=0.115\textwidth]{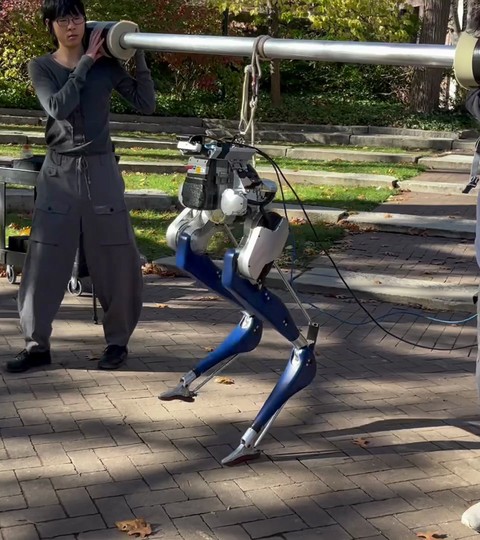}
		\includegraphics[width=0.115\textwidth]{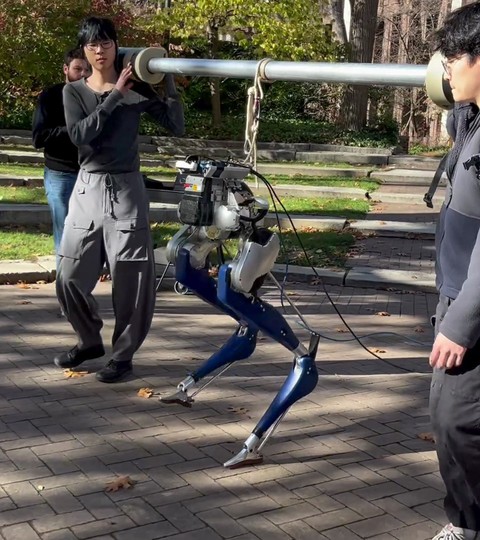}\\
		\vspace{3pt}
		\includegraphics[width=0.115\textwidth]{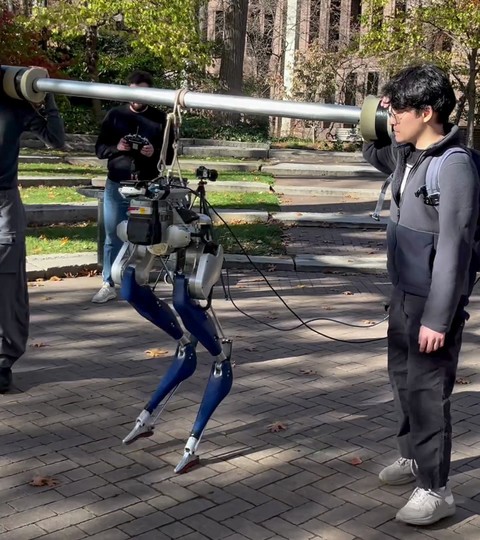}
		\includegraphics[width=0.115\textwidth]{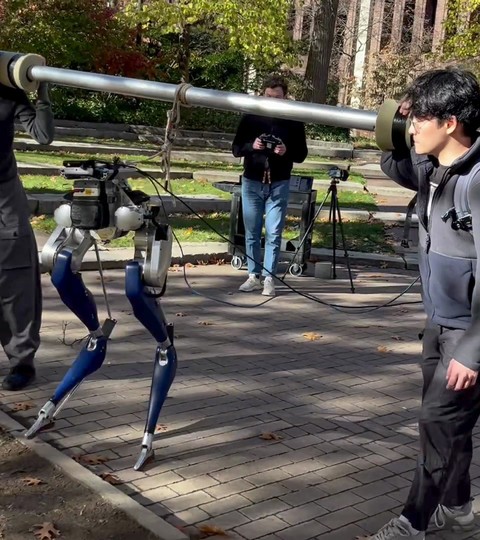}
		\includegraphics[width=0.115\textwidth]{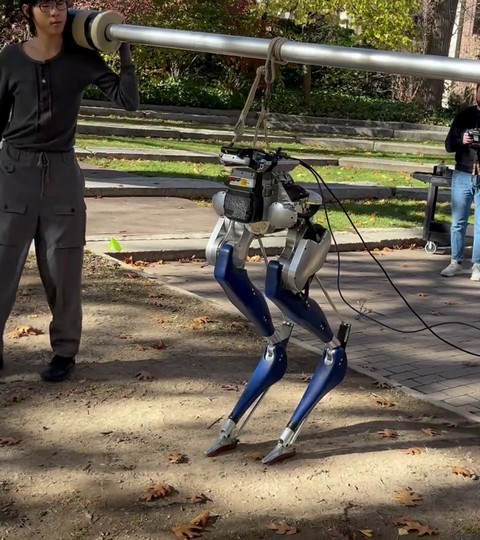}
		\includegraphics[width=0.115\textwidth]{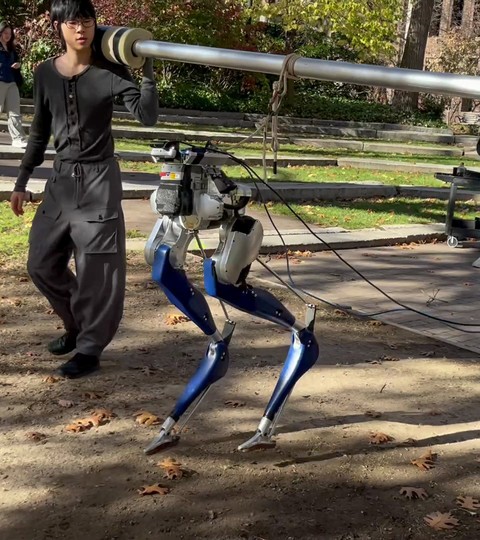}
		\includegraphics[width=0.115\textwidth]{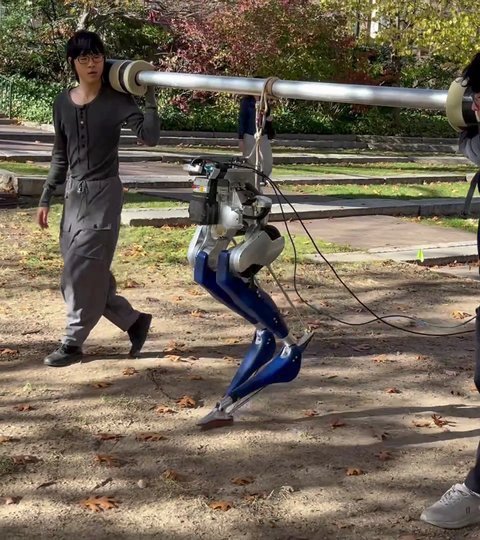}
		\includegraphics[width=0.115\textwidth]{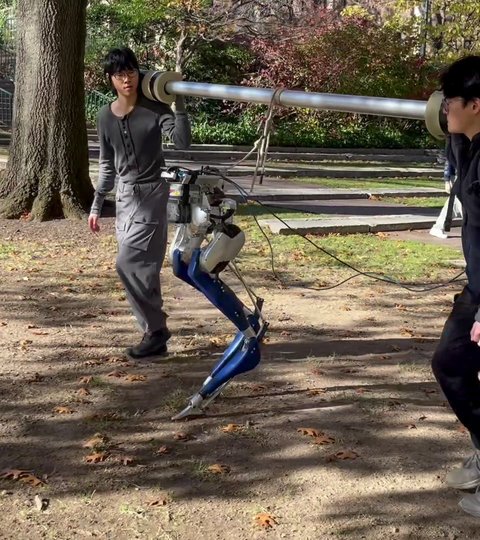}
		\includegraphics[width=0.115\textwidth]{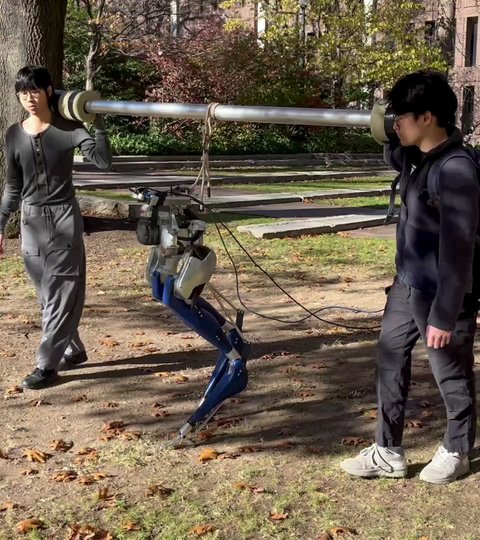}
		\includegraphics[width=0.115\textwidth]{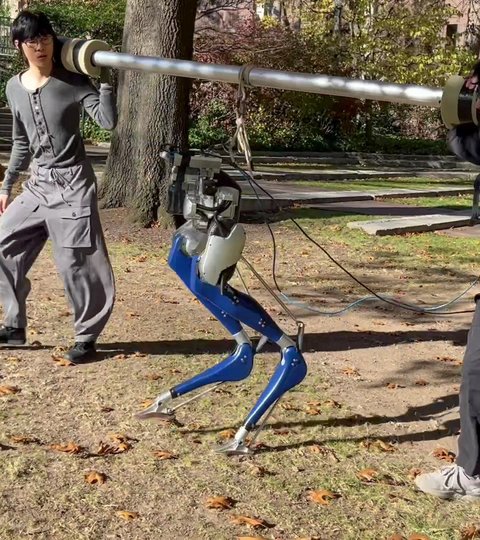}\\
		\vspace{3pt}
		\includegraphics[width=0.115\textwidth]{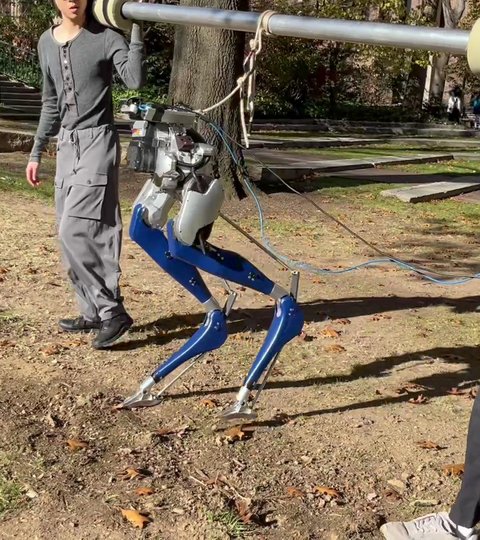}
		\includegraphics[width=0.115\textwidth]{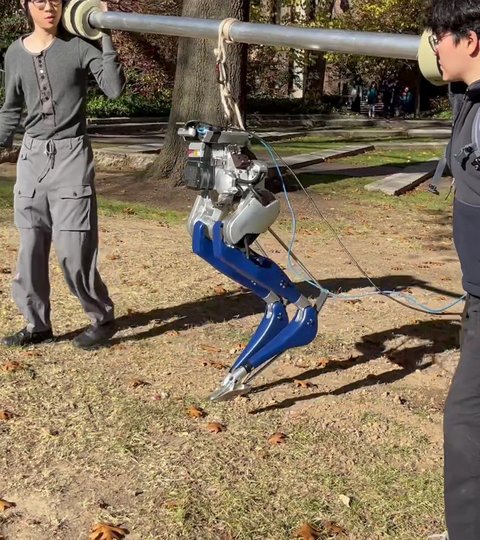}
		\includegraphics[width=0.115\textwidth]{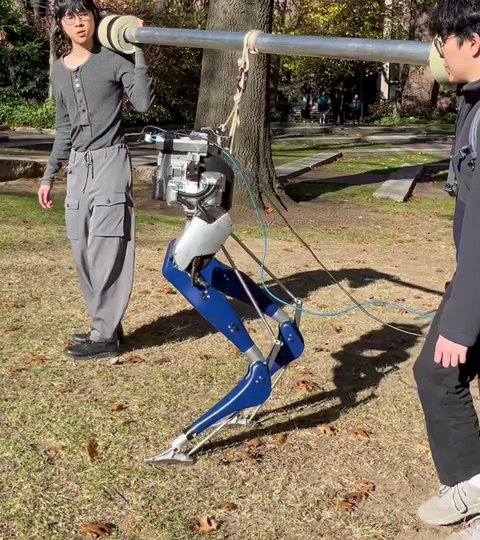}
		\includegraphics[width=0.115\textwidth]{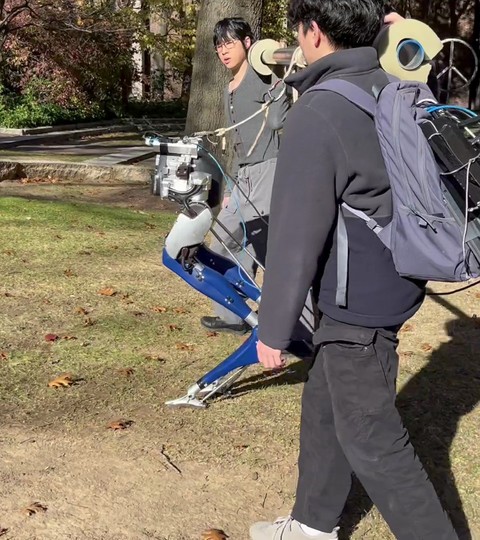}
		\includegraphics[width=0.115\textwidth]{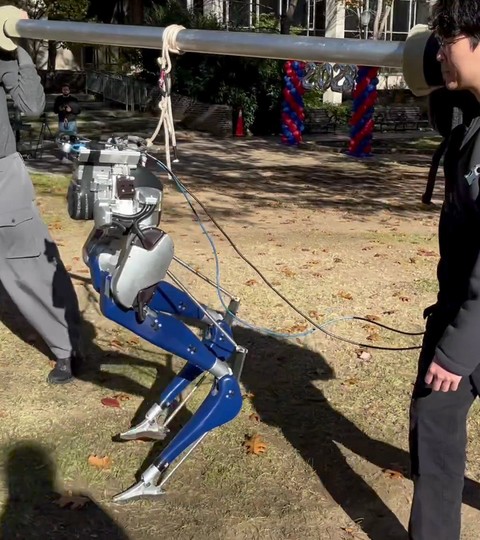}
		\includegraphics[width=0.115\textwidth]{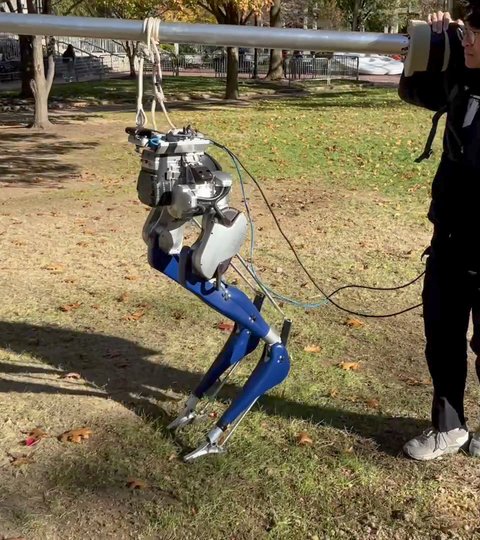}
		\includegraphics[width=0.115\textwidth]{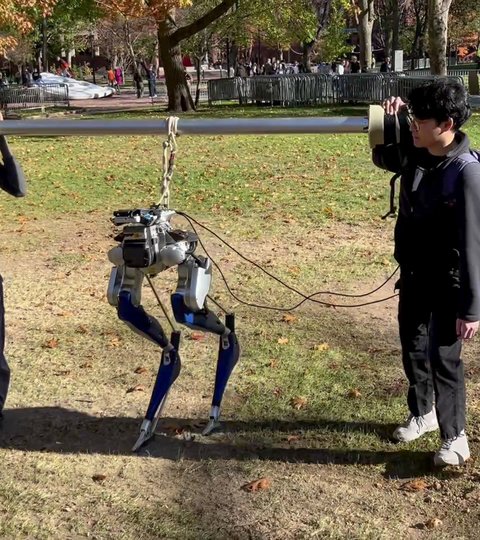}
		\includegraphics[width=0.115\textwidth]{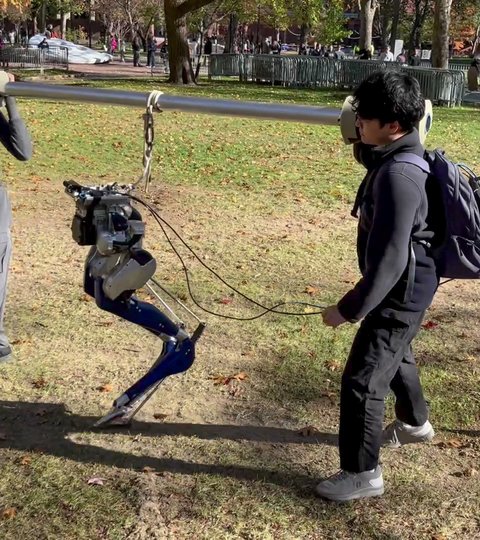}
		\caption{Motion tiles showing Cassie ascending and descending steps, stepping over a curb onto the grass, and walking up a grassy slope in one continuous walking trial. \vspace{1em}}
		\label{fig:stairs-and-grass-motion-tiles}
	\end{subfigure}
	\vspace{1em}
	\begin{subfigure}[t]{0.32\textwidth}
		\centering
		\includegraphics[width=0.99\textwidth]{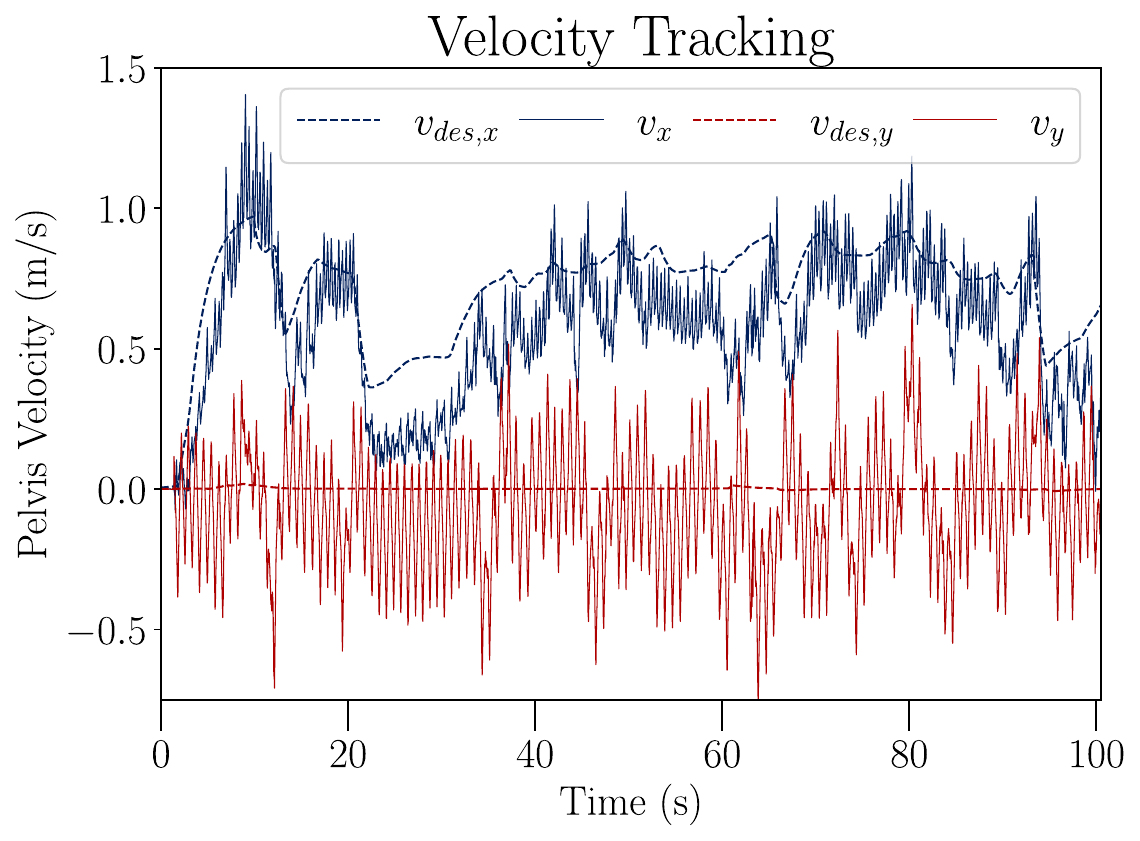}
		\caption{Plot of the velocity tracking performance of the robot using our control stack.}
		\label{fig:stairs-and-grass-vel-tracking}
	\end{subfigure}
	~
	\begin{subfigure}[t]{0.32\textwidth}
		\centering
		\includegraphics[width=0.99\textwidth]{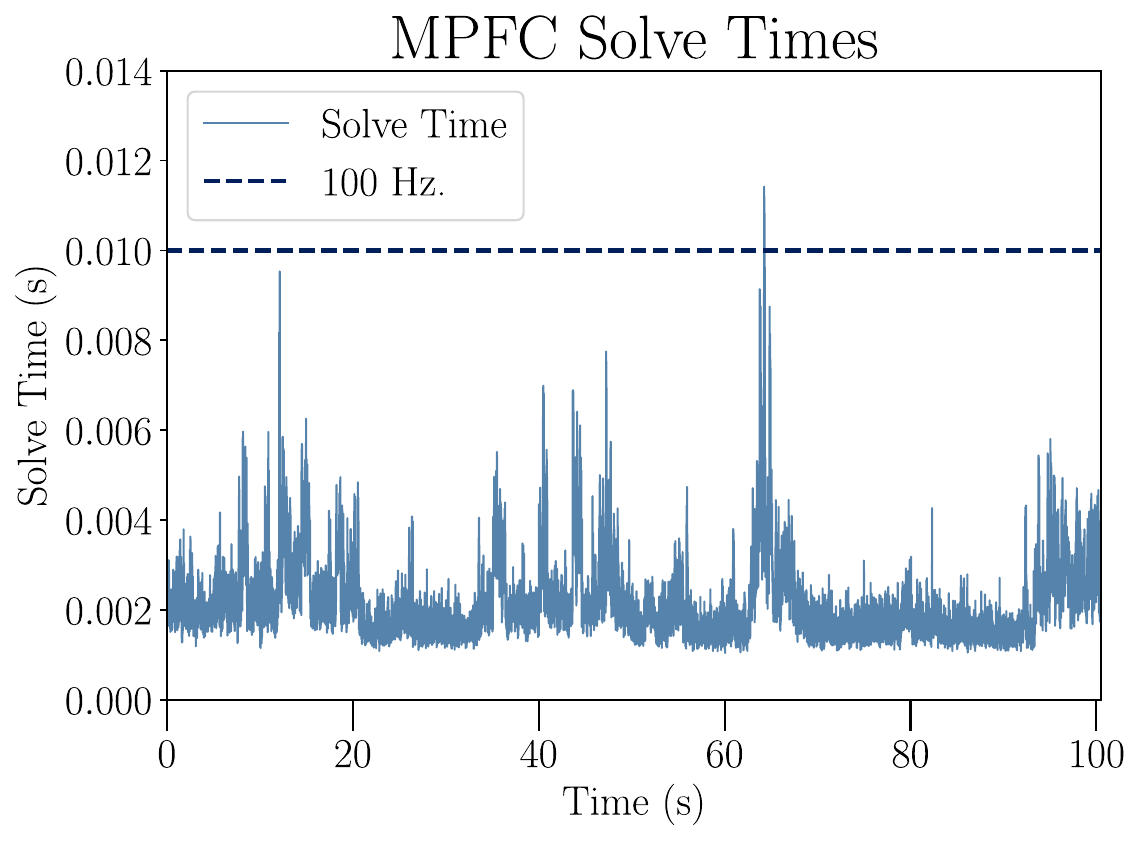}
		\caption{MPFC Solve times during the above trial.}
		\label{fig:stairs-and-grass-solve-time}
	\end{subfigure}
	~
	\begin{subfigure}[t]{0.32\textwidth}
		\centering
		\includegraphics[width=0.99\textwidth]{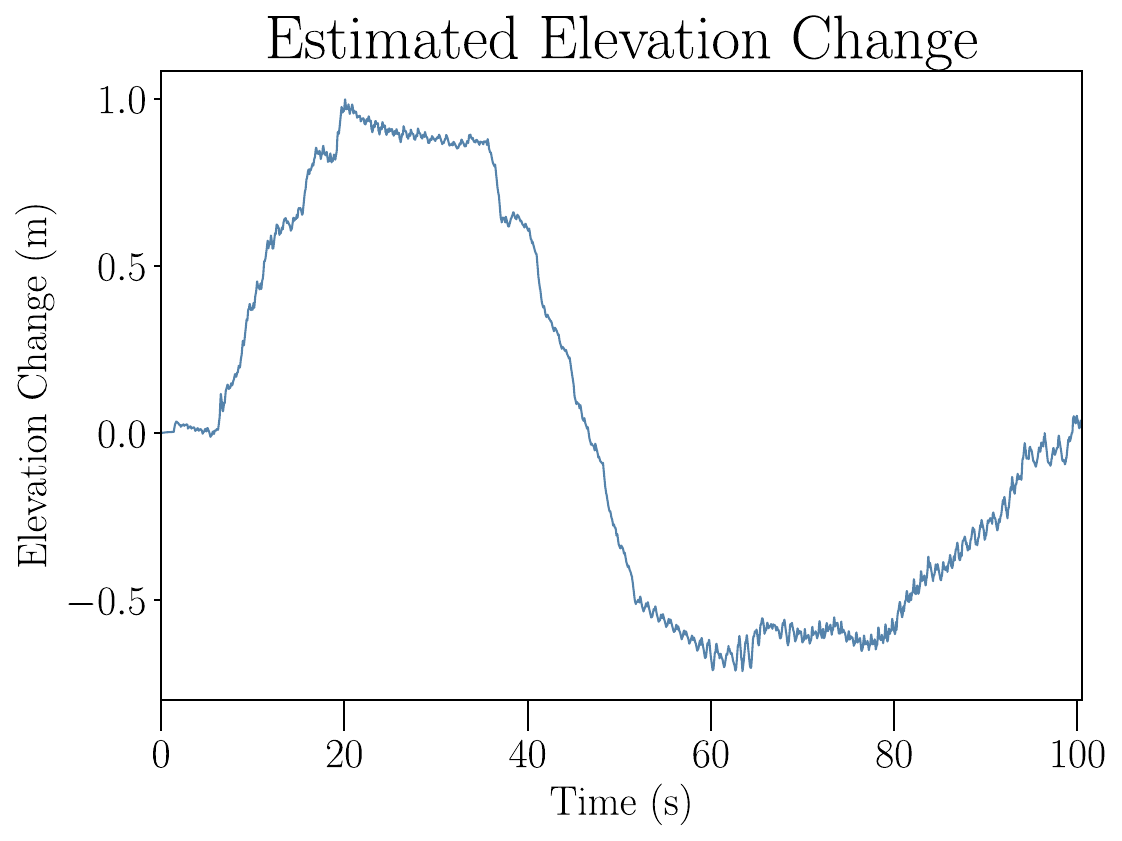}
		\caption{Plot of the elevation change over the trial, estimated from the onboard state estimator. }
		\label{fig:stairs-and-grass-pelvis_height}
	\end{subfigure}
	\caption{Cassie Walks on unstructured terrain using our proposed perception and control stack, climbing and descending a set of steps, stepping over a curb, and walking up a grassy hill. 
		Our perception stack identifies safe terrain and decomposes it into convex polygons online while the robot is walking at over 0.5 m/s.
		Footage can be viewed in the supplemental video.}
	\label{fig:results_hero}
\end{figure*}

%% file: chapters/discussion.tex
\section{Discussion}
This section discusses implementation details, limitations, and failure modes.
\subsection{Implementation Details}
First, we discuss design choices introduced to handle edge cases and increase the robustness of our implementation. 

\subsubsection{Inpainting}
Because we only use a single depth camera, whose field of view does not span the entire diagonal of the elevation map, we lack elevation data for terrain near the robot when not walking straight forward, leaving the question of how S3 should classify these cells. 
During our hardware testing, classifying these cells as unsafe caused the robot to fall when the operator drove the robot toward unmapped regions. 
We solve this by inpainting the missing portions of the elevation map using the Navier-Stokes based method implemented in OpenCV~\cite{bertalmioInpaintNS}, before inputting the elevation map to S3.
This method matches the value and gradient of the image at the boundary of the missing terrain. 
\brian{For many real world terrains, this continuous extrapolation is a safe assumption, since obstacles extend uni-directionally across the entire map. 
	In more dangerous environments, and when missing elevation map values are primarily due to occlusions rather than a lack of sensor coverage, such as in our simulation stepping stone experiments, a more conservative inpainting scheme such as least-neighboring-value \cite{mikiElevationMappingLocomotion2022} is appropriate.}
The ideal solution would include additional depth sensors, however our solution highlights a general theme: due to Cassie's underactuation, it is often safer to resolve \textit{ambiguous} design decisions by favoring steppability. 

\subsubsection{ALIP State Estimation}
Impacts during touchdown and compliance in Cassie's hip-roll joints can cause undesirable spikes and oscillations in the lateral floating base velocity estimate, and therefore the angular momentum estimate. 
We increase our controller's robustness to these issues by using a Kalman filter with ALIP dynamics to smooth our estimate of the ALIP state during single support. We use \eqref{eq:alip} for the dynamics model, with full-state measurement, and assume a much higher measurement noise for the angular momentum than for the CoM position. 

\subsubsection{Foostep Height Lookup}
\begin{figure}
	\centering
	\includegraphics[width=\linewidth]{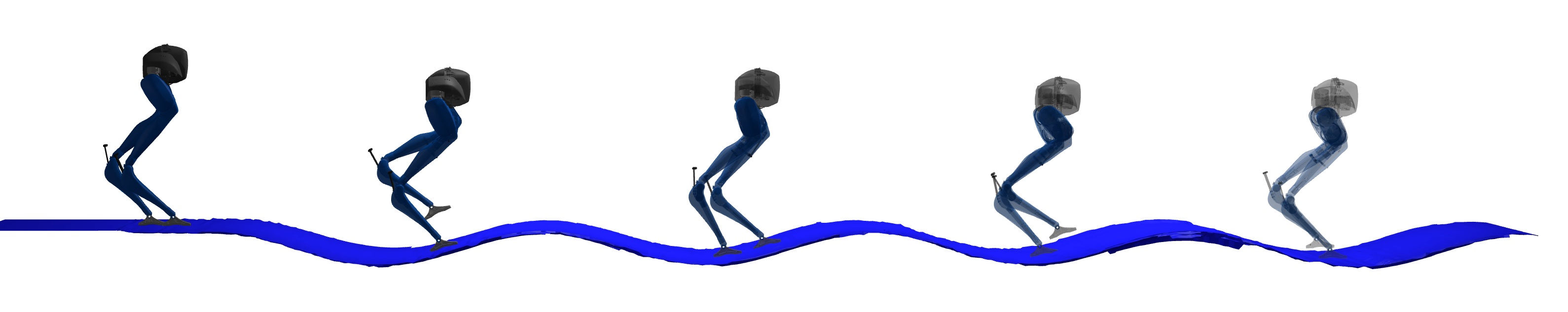}
	\caption{\brian{Our proposed system naturally handles terrain with no obvious planar approximation. 
		Despite constantly varying height and surface normals, S3 classifies the entirety of this sinusoidal terrain as safe, resulting in a single steppable region matching the extents of the map. 
		Because ALIP dynamics are height-independent, the z-coordinate of each planned footstep does not affect the rest of the MPC solution. 
		Therefore, we use the elevation map to determine the footstep height sent to the low-level OSC.}}
	\label{fig:undulation}
\end{figure}

Before sending a footstep command to the OSC, we refine the vertical footstep position by looking up the height of the planned footstep position on a smoothed, inpainted copy of the elevation map. 
Because the ALIP dynamics do not depend on the vertical footstep position, we do not need to propogate this adjustment back to MPFC. 
\brian{In addition to increasing the practical robustness of the system, this allows Cassie to walk on undulating terrain without any modifications to the perception or control stack (\cref{fig:undulation}).}

\subsection{Limitations, Failure Modes, and Future Work}
\label{subsec:limits}
This section discusses limitations and failure modes, providing directions for future research. 
\brian{We organize these into systems limitations, algorithmic limitations, and fundamental limitations. These limitations would respectively require engineering effort, further research into the proposed methods, or structural changes to the approach to resolve.}
\subsubsection{Systems Limitations}
\brian{We use single threaded CPU implementations of elevation mapping and S3, limiting the map size and resolution which can be handled in real time.
Existing GPU-based elevation mapping implementations \cite{mikiElevationMappingLocomotion2022} could be used with a GPU implementation of S3 to handle larger or more detailed maps. 
Additionally, the fast swing foot motions and CoM height changes required to walk on steps pushed the boundaries of what could be tracked with our OSC, limiting the step heights traversable on hardware to 16 cm. 
More challenging terrains will require considering swing-foot and vertical CoM dynamics at the MPC level, either by incorporating more detailed dynamics into MPFC, or by using whole-body MPC to realize the MPFC footstep plans. 

\subsubsection{Algorithmic Limitations}
The planning horizon of 2 footsteps can result in overly optimistic footstep choices.
Further research could focus on stronger mixed integer formulations or improved mixed-integer solvers to enable solving for longer footstep horizons in real time. 
Alternatively, additional robustness terms could be incorporated in MPFC to favor conservative behaviors. 
Cassie's small lateral workspace and MPFC's fixed stepping pattern make the controller vulnerable to lateral perturbations, especially those occurring  at the beginning of single-stance. 
Robustness against these perturbations could be increased by including crossover steps, and by smaller minimum stance times. 
The stance duration should then be coupled to the swing-foot workspace to maintain reachability of the planned footsteps. 

This paper uses heuristic steppability criteria, which demonstrate the advantages of S3 compared to plane segmentation, but do not investigate other potential benefits of S3 as a general framework.
For example, one failure case involved dried leaves which had accumulated underneath the edge of a step, filling in the space and resulting in the edge being classified as steppable.
This could be avoided by incorporating a higher level semantic segmentation as an additional safety criterion.}

\subsubsection{Fundamental Limitations}
The failure mode experienced on hardware which was most insufficiently addressed by this work was slipping. 
Most often, the robot would fall immediately upon slipping, but if it recovered, the slip could introduce large errors into the elevation map which lead failure \brian{from an incorrect segmentation or error in the estimated ground height.}
The likelihood of slips could be reduced by more conservatively constraining the workspace of the ALIP model (effectively the friction cone), \brian{but this cannot entirely eliminate the possibility. 
This vulnerability highlights that like all model-based approaches, ours is vulnerable to gaps between modeling assumptions and the real world.
S3 prevents inconsistent segmentation of reasonable elevation maps from causing failures, but cannot correct maps which inaccurately reflect the real environment.
For situations such as tall grass, or recovering from slip-induced errors, methods are needed which can adaptively rely on either perception or proprioception, and recognize and reset invalid maps. 
}

%% file: chapters/conclusion.tex
\section{Conclusion And Future Work}\label{sec:conclusion}
We present a complete perception and control stack for underactuated bipedal walking on rough terrain. 
We formulate Model Predictive Footstep Control as a single MIQP which can be solved at over 100 Hz. to stabilize walking over discontinuous terrain without a pre-specified foothold sequence. 
Motivated by the brittleness of plane segmentation for safe terrain classification, we develop Stable Steppability Segmentation, a simple algorithm for temporally consistent safe terrain segmentation, and a complementary convex polygon decomposition algorithm for generating foothold constraints online. 
We demonstrated our proposed perception and control stack on the underactuated Cassie biped through outdoor experiments. 
Future work will consider more expressive models than the ALIP, to increase the robustness of the controller and allow bipeds to walk on shallow footholds and execute large step-to-step height changes.

%% file: chapters/appendix.tex
\subsection{Lateral Reset Map Adjustment}
\label{sec:reset_appendix}

Because $B_{ds}$ is decoupled in $x$ and $y$, our \brian{hardware} MPFC implementation uses $f(t) = 1$ for the lateral ALIP state components, corresponding to instantaneous weight transfer on touchdown.
In this case, \eqref{eq:b_ds_general} evaluates to 
\begin{equation}
	B_{ds} = A^{-1}(A_{r} - I)B_{CoP}. 
	\label{eq:b_ds_1}
\end{equation}
This helps realize the desired step width by compensating for systematic error in swing foot tracking. 
The robot consistently steps wider than the commanded footstep position. 
Due to compliance and backlash in Cassie's hip roll joints, we cannot increase the swing-foot PD gains beyond the values in \cref{tab:osc_params_hw}. 
Assuming instantaneous lateral weight transfer acts as a feed-forward correction by increasing the model's estimate of how much momentum will be absorbed by a larger lateral footstep size. 
To calculate the final value of $B_{ds}$, we take the inner $2\times2$ submatrix, which corresponds to the coronal plane, from \eqref{eq:b_ds_1}, and the 4 corner values from \eqref{eq:b_ds_t}, which correspond to the sagittal plane. 

\subsection{Constructing the Desired-Velocity Subspace}
\label{sec:subspace}
\brian{Here we show how to derive \eqref{eq:subspaces_for_cost}. 
For this analysis, we ignore the $z$ component of each footstep, since it does not enter the ALIP dynamics, and assume that $p_{n} \in \mathbb{R}^{2}$.}
We will define the projection matrices $\Pi_{0}$ and $\Pi_{1}$, and the offsets $d_{0}$ and $d_{1}$, and then for the general case, we have that 
\begin{align}
	\Pi_{n+2} = \Pi_{n} \nonumber \\
	d_{n+2} = d_{n} \nonumber
\end{align}

\brian{We start by substituting \eqref{eq:subspace_2} into \eqref{eq:subspace_1} and solving for $x_{0}$}:

\begin{equation}
	x_{0} = G(A_{s2s}B_{s2s} - B_{s2s})\delta p_{0} + 2 T_{s2s}GB_{s2s}v_{des}
	\label{eq:subspace_constructive}
\end{equation}

where $G = (I - A_{s2s}^{2})^{-1}$. 
\eqref{eq:subspace_constructive} defines the desired velocity subspace as an offset based on $v_{des}$ and the span of $L_{0} = G(A_{s2s} - I)B_{s2s} \in \mathbb{R}^{4\times2}$. 
We convert this to the desired form \eqref{eq:subspaces_for_cost} by left-multiplying with $\Pi_{0}$, a projection matrix to the orthogonal complement of the range of $L_{0}$. 
Because $\Pi_{0}$ maps $L_{0} \delta p_{0}$ to zero for any $\delta p_{0}$ by construction, this leaves us with

\begin{equation}
\Pi_{0}x_{0} = 2 \Pi_{0}T_{s2s} G B_{s2s}v_{des}
\end{equation}

so  $d_{0} (v_{des}) = 2 T_{s2s}GB_{s2s} v_{des}$. To find $\Pi_{1}$, we use 

\begin{align}
	& x_{1} = A_{s2s} x_{0} + B_{s2s} \delta p_{0} \nonumber \\
	\therefore \text{ }& x_{1} = A_{s2s}(L_{0} \delta p_{0} + d_{0}) + B_{s2s} \delta p_{0} \nonumber\\
	\therefore \text{ } & x_{1} = (A_{s2s}L_{0} + B) \delta p_{0} + A_{s2s}d_{0} \nonumber \\
	\therefore \text{ } & L_{1} =  A_{s2s}L_{0} + B, d_{1} = A_{s2s} d_{0}
\end{align}

And $\Pi_{1}$ is similarly constructed as a projection to $\text{span}(L_{1})^{\perp}$.

\subsection{Whittling Algorithm Cut Solver}
\label{sec:whittling_solver}
 This section presents a solver for optimization over $S^{1}$, which we use to solve \eqref{eq:make_cut} quickly online. 
 Given an optimization problem 
 
\begin{align*}
	\label{eq:optim_on_s1}
	\underset{x \in \mathbb{R}^2}{\text{minimize }} & f(x) \\
	\text{subject to } & \lVert x \rVert_{2}^{2} = 1, 
\end{align*}
the associated first-order optimality conditions are 
\begin{align}
	\lVert x \rVert_{2}^{2} = 1\\
	\nabla f(x) + \nu x = 0
\end{align}

where $\nu \in \mathbb{R}$ is a Lagrange multiplier for the unit-norm constraint. 
To optimize over the unit circle, we rotate $x$ in the direction which decreases the cost until $\nabla f(x)$ is parallel to $x$, which satisfies the optimality conditions. 
Our solver is summarized in \cref{alg:whittling_solver}.
 
 \begin{algorithm}[H]
 	\caption{MakeCut Solver} \label{alg:whittling_solver}
 	\begin{algorithmic}[0]
 		\Require Cost function $f$, Initial guess $x \in S^{1}$, Optimality Tolerance $\epsilon$, Line search parameters $\alpha > 0, \beta \in (0, 1)$
 		\Procedure{Solve}{$f$, $x$, $\epsilon$}
 		\State$\theta \gets \infty$
 		\While{$\vert \theta \vert > \epsilon$}
 		\State $\theta \gets (\nabla f(x) - x\langle \nabla f(x), x \rangle) \times x $
 		\State $t \gets \alpha / \vert \theta \vert$
 		\While{$f(\text{Rotate}(\theta t, x)) > f(x)$} $t \gets \beta t$
 		\EndWhile
 		\State $x \gets \text{Rotate}(\theta t, x)$
 		\EndWhile
 		\State \Return $x$ 
 		\EndProcedure
 	\end{algorithmic}
 \end{algorithm}
 
where 
\begin{align*}
	\text{Rotate}(\theta, x) = \begin{bmatrix} 
		\cos \theta & -\sin \theta \\
		\sin \theta & \cos \theta
	\end{bmatrix} x.
\end{align*}

The $\theta$ update finds the direction to rotate $x$ by considering the component of $\nabla f$ orthogonal to $x$, using the cross product with $x$ to convert this direction into a scalar rotation angle. 
We then perform a line search, starting with a fixed initial step size $\alpha$ for improved convergence speed. 
As an implementation note, we re-normalize $x$ at each iteration to avoid drift in the unit-norm constraint.

\subsection{Controller and Perception Stack Parameters}
\label{sec:params_apendix}
The following tables give the parameters used for each component of our stack. Diagonal matrices are represented as $\text{d}[\cdots ]$, where the arguments to $\text{d}$ represent the entries on the diagonal of the matrix.
 \brian{While we did not extensively tune MPFC costs, we found it worked best to set $Q_{N}$ at least 100$\times$ larger than $Q$ for the position coordinates. 
 This avoided short-sighted behavior when walking over edges, and could maybe be reduced for longer planning horizons.
 
 Compared to hardware, our simulation experiments feature increased MPFC state costs and OSC PD gains, and decreased S3 resolution and hysteresis.
 These differences showcase the full potential of our method with more-precise swing-foot tracking. 
 We switch the inpainting approach from Navier-Stokes (NS) to Least-Neighboring-Value (LNV) to align with the discrete terrain tested in sim.
}
\begin{table}[h]
	\centering
	\caption{MPFC Parameters}
	\begin{tabular}{clll}
		\hline
		\textbf{Symbol} & \textbf{Meaning} & \textbf{Hardware} & \textbf{Simulation} \\
		\hline
		$N$ & MPFC Horizon & 2 steps & 2 steps\\
		$t_{min}$ &  Min. SS duration & 0.27 s & 0.27 s \\
		$t_{max}$ &  Max. SS duration & 0.33 s & 0.33 s \\
		$H$ & ALIP height & 0.85 m & 0.85 m\\
		$T_{ss}$ &  Nominal SS duration & 0.3 s & 0.3 s\\
		$ T_{ds}$ & DS duration &  0.1 s & 0.1 s\\
		$ w_{T}$ & Time weight & 100 & 100\\
		$l$ & Step width &  0.2 m & 0.15 m\\
		$ w_{u}$ & Ankle torque weight & 0.01 & 0.01 \\
		$u_{max}$ & Max. ankle torque & 22 Nm & 22 Nm\\
		$Q_{N}$ & Terminal state cost & \makecell[l]{$\text{d}[100, 100,$\\ \qquad \quad$ 1, 1]$} & \makecell[l]{$\text{d}[1000, 1000,$\\ \ \qquad \quad $ 20, 20]$}\\
		$Q$ & Runnning state cost & \makecell[l]{$\text{d}[0.001, 0.1,$\\ \quad $0.01, 0.001]$} & $\text{d}[10, 10, 5, 5]$\\
		$R$ & Running step size cost & $\text{d}[25, 25, 0]$ & $\text{d}[25, 25, 0]$ \\
		-- & CoM soft pos. limits &$\pm$ [0.35, 0.35] m & $\pm$ [0.4, 0.4] m\\
		-- & CoM soft vel. limits & $\pm$ [2.5, 1.5] m/s &  $\pm$ [2.5, 1.5] m/s \\
		-- & Soft constraint cost & 1000 & 1000\\
		\hline
	\end{tabular}
\end{table}
\vspace{-0.5cm}

\begin{table}[h]
	\centering
	\caption{OSC Gains (Hardware)}
	\begin{tabular}{llll}
		\hline
		\textbf{OSC Objective} & \textbf{W} & \textbf{Kp} & \textbf{Kd} \\
		\hline
		Toe joint angle & 1 & 1500 & 10 \\
		Hip yaw angle & 2 & 40 & 2 \\
		CoM [x, y, z] & [0, 0, 10] & [0, 0, 100] & [0, 0, 6] \\
		Pelvis [roll, pitch, yaw] & [2, 4, 0.02] & [200, 200, 0] & [10, 10, 4] \\
		Swing Foot [x, y, z] & [4, 4, 2] & [220, 180, 180] & [6, 5.5, 5.5] \\
		Ankle Torque & 10 & -- & -- \\
		\hline
	\end{tabular}
	\label{tab:osc_params_hw}
\end{table}
\vspace{-0.5cm}

\begin{table}[h]
	\centering
	\caption{OSC Gains (Simulation)}
	\begin{tabular}{llll}
		\hline
		\textbf{OSC Objective} & \textbf{W} & \textbf{Kp} & \textbf{Kd} \\
		\hline
		Toe joint angle & 1 & 1500 & 10 \\
		Hip yaw angle & 2 & 100 & 4 \\
		CoM [x, y, z] & [0, 0, 10] & [0, 0, 80] & [0, 0, 10] \\
		Pelvis [roll, pitch, yaw] & [2, 4, 0.02] & [200, 200, 0] & [10, 10, 10] \\
		Swing Foot [x, y, z] & [4, 4, 2] & [400, 400, 400] & [20, 20, 25] \\
		Ankle Torque & 10 & -- & -- \\
		\hline
	\end{tabular}
	\label{tab:osc_params_sim}
\end{table}

\vspace{-0.5cm}
\begin{table}[h]
	\centering
	\caption{Perception Stack Parameters}
	\begin{tabular}{clll}
		\hline
		\textbf{Symbol} & \textbf{Meaning} & \textbf{Hardware} & \textbf{Simulation} \\
		\hline
		\multicolumn{3}{c}{\textit{Elevation Mapping}} & \\
		-- & Map Size & $3 \times 3$ m & $2.5 \times 2.5$ m\\
		-- & Map Resolution & $0.03$ m & $0.025$ m \\
		\hline
		\multicolumn{3}{c}{\textit{S3}} & \\
		$k_{hyst}$ & Safety Hysteresis & 0.6 & 0.4\\
		$k_{safe}$ & Safety Threshold & 0.7 & 0.7\\
		-- & Safety Margin Kernel Size & 4 px & 4 px \\
		$\sigma_{LoG}$ & LoG Standard Dev. for $c_{curve}$ & 2 px & 2 px\\
		$\alpha_{c}$ & $c_{curve}$ scaling parameter & 5 & 5\\
		-- & $c_{inc}$ kernel size & 5 px & 5 px\\
		-- & Inpainting & NS & LNV\\
		\hline
		\multicolumn{3}{c}{\textit{Convex Decomposition}} & \\
		$d$ & ACD Concavity Limit & 0.25 m & 0.25 m\\
		\hline
	\end{tabular}
\end{table}